\documentclass[lettersize,journal]{IEEEtran}
\usepackage{amsmath,amsfonts}
\usepackage{algorithmic}
\usepackage{algorithm}
\usepackage{array}
\usepackage[caption=false,font=scriptsize,labelfont=rm,textfont=rm]{subfig}
\usepackage{textcomp}
\usepackage{stfloats}
\usepackage{url}
\usepackage{verbatim}
\usepackage{graphicx}
\usepackage{cite}
\usepackage{enumitem}
\usepackage{booktabs} 
\usepackage{etoolbox}
\usepackage{multirow}
\usepackage{microtype}
\usepackage{orcidlink}
\hypersetup{hidelinks,colorlinks=true,allcolors=black,pdfstartview=Fit,breaklinks=true} 
\makeatletter
\patchcmd{\@makecaption}
{\scshape}
{}
{}
{}
\makeatother

\usepackage{colortbl}
\begin{document}

\title{Downstream-Pretext Domain Knowledge Traceback for Active Learning}

\author{Beichen~Zhang, Liang~Li, Zheng{-}Jun~Zha, Jiebo~Luo,~\IEEEmembership{Fellow,~IEEE} and 
	Qingming~Huang,~\IEEEmembership{Fellow,~IEEE} 

	\IEEEcompsocitemizethanks{
		\IEEEcompsocthanksitem B. Zhang and Q. Huang are with  the School of Computer Science and Technology, University of Chinese Academy of Sciences, Beijing 101408, China(E-mail: zhangbeichen14@mails.ucas.ac.cn; qmhuang@ucas.ac.cn). 
		\IEEEcompsocthanksitem L. Li is with the Key Laboratory of Intelligent Information Processing, Institute of Computing Technology, Chinese Academy of Sciences, Beijing 100190, China (E-mail: liang.li@ict.ac.cn).(\emph{Corresponding author: Liang Li.})
		\IEEEcompsocthanksitem Z. Zha is with the School of Information Science and Technology, University of Science and Technology of China, Hefei 230027, China (E-mail: zhazj@ustc.edu.cn).
		\IEEEcompsocthanksitem J. Luo is  with the Department of Computer Science, University of Rochester, Rochester, NY 14627 USA (E-mail: jluo@cs.rochester.edu).
		
}}

\markboth{IEEE TRANSACTIONS ON MULTIMEDIA, 2024}%
{Shell \MakeLowercase{\textit{et al.}}: A Sample Article Using IEEEtran.cls for IEEE Journals}


\maketitle

\begin{abstract}
Active learning (AL) is designed to construct a high-quality labeled dataset by iteratively selecting the most informative samples.
Such sampling heavily relies on data representation, while recently  pre-training is popular for  robust feature learning.  
However, as pre-training utilizes low-level pretext tasks that lack annotation, directly using pre-trained representation in AL is inadequate for determining the sampling score.
To address this problem, we propose a downstream-pretext domain knowledge traceback (DOKT) method that traces the data interactions of downstream knowledge and pre-training guidance for selecting diverse and instructive samples near the decision boundary. 
DOKT consists of a traceback diversity indicator and a domain-based uncertainty estimator.
The diversity indicator constructs two feature spaces based on the pre-training pretext model and the downstream knowledge from annotation, by which it locates the neighbors of unlabeled data from the downstream space in the pretext space to explore the interaction of samples.
With this mechanism, DOKT unifies the data relations of low-level and high-level representations to estimate traceback diversity.
Next, in the uncertainty estimator, domain mixing is designed to enforce perceptual perturbing to unlabeled samples with similar visual patches in the pretext space.
Then the divergence of perturbed samples is measured to estimate the domain uncertainty.
As a result, DOKT selects the most diverse and important samples based on these two modules.
The experiments conducted on ten datasets show that our model outperforms other state-of-the-art methods and generalizes well to  various application scenarios such as semantic segmentation and image captioning.
\end{abstract}


\section{Introduction}
\label{sec:1}
The success of deep neural networks relies heavily on the use of large-scale labeled data to train the associated models to achieve good performance.
In view of this paradigm, active learning (AL)~\cite{settles2009active} was proposed for constructing high-quality labeled datasets by labeling only the most representative samples while maximizing the training performance.
Since AL methods commonly estimate the importance of samples by analyzing their underlying data distribution, the performance of AL-based sampling heavily depends on the quality of the given data representation.
The previous AL methods learn representations based on labeled/unlabeled data, which is an inefficient approach due to the presence of inadequate annotations in the early sampling stages and the limited scale of the input data.
Recently, self-supervised pre-trained models~\cite{chen2020simple,bao2021beit, he2022masked, 9362305} learn high-quality representations via large-scale pretext tasks.
However, the representations in pretext tasks do not involve domain knowledge with manual annotations; therefore, the AL method cannot select instructive samples near the decision boundary.
Therefore, this paper explores how to embed downstream domain knowledge into a pre-training feature space for efficiently implementing AL sampling.

\begin{figure}[t]
	\begin{center}

		\includegraphics[width=0.81\linewidth,trim=10 60 90 40]{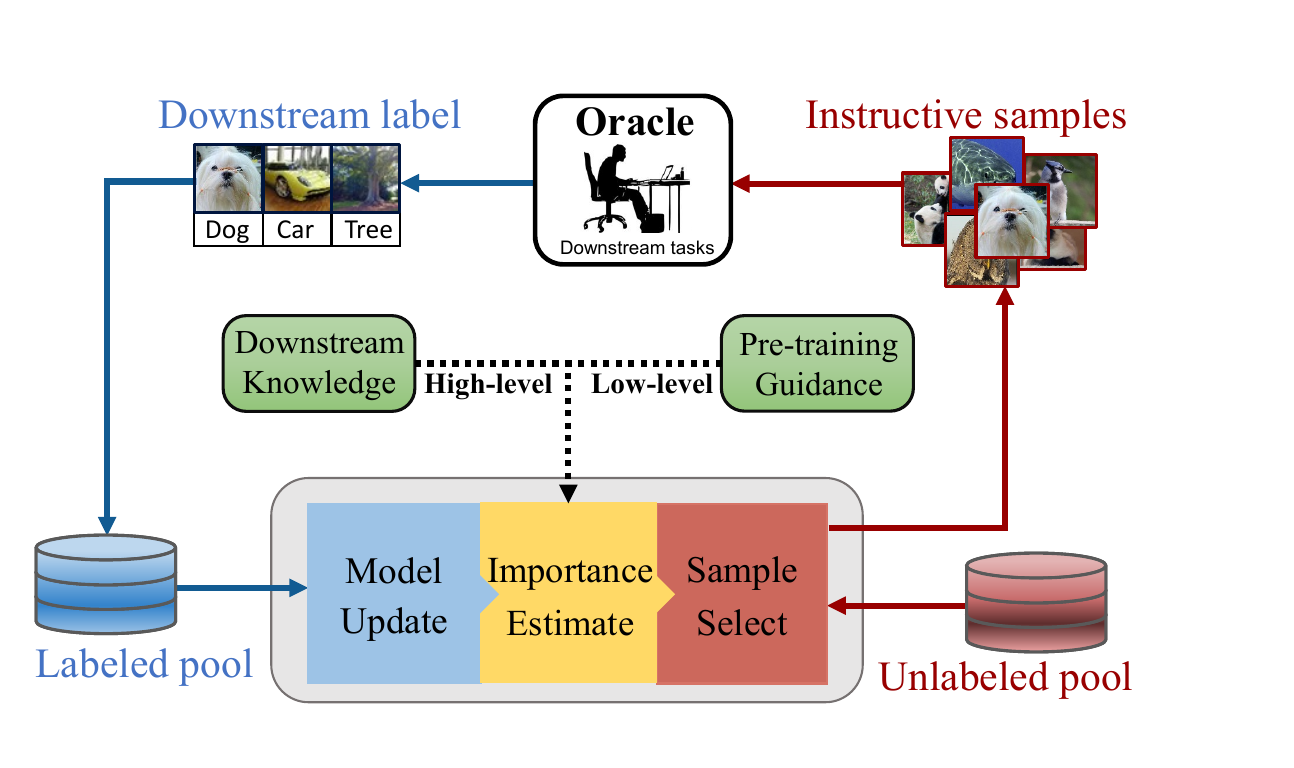}
	\end{center}
	\caption{A pool-based AL cycle.  In each iteration, the AL model is trained with labeled data. After training, a subset of unlabeled samples is selected based on the model inference and then labeled by an oracle. We leverage pre-training guidance and downstream knowledge to evaluate sampling score for better AL sampling. This AL cycle repeats until the model performance meets the user's requirements or the label budget runs out.}
	\label{img1}
\end{figure}

As shown in Fig.~\ref{img1}, an AL algorithm is typically an iterative schedule in which a set of samples are selected from an unlabeled pool to be labeled in each iteration.
Its goal is to maximize the training performance with a limited annotation budget.
Currently, the popular AL methods can be categorized as distribution-based or uncertainty-based methods.
Distribution-based methods choose samples that are diverse for covering the distribution of the whole dataset.
Uncertainty-based AL methods label the most uncertain samples to improve the target model.
In both strategies, data representations embedded with key information serve to estimate the data distribution or quantify data uncertainty, and this step plays an essential role.
However, since AL iteratively trains a model from scratch, the available labeled samples are insufficient for supporting robust representation learning, particularly in early AL iterations.
Without reliable data representations, distribution-based methods cannot capture data relationships to estimate data diversity, and uncertainty-based methods cannot locate the decision boundary to select the most uncertain samples.
This results in unreliable data diversity and uncertainty estimates.
Furthermore, the limited data scale and long-tailed label distribution also introduce serious bias to the AL.

Recently, self-supervised pre-trained models are proposed to learn effective representation from large-scale datasets. 
With the help of pretext tasks, the pre-trained model is perceptive to visual information and  provides high-quality data representation, which can help to reduce the above shortcoming for AL.
However, as pre-training only learns from low-level pretext tasks without annotation, these model has good visual perception but lacks domain knowledge for manual annotation. 
Directly using pre-trained representations can only help to select samples with diverse visual patterns rather than key information of high-level semantics.
Thus, directly using pre-training representations of pretext tasks is incomplete to model the relationship between labeled and unlabeled data, and cannot determine the most important samples for constructing a high-quality dataset and maximizing model training.




In this paper, we propose a \textbf{d}ownstream pretext-d\textbf{o}main \textbf{k}nowledge \textbf{t}raceback (DOKT) method for AL that traces data interactions between pretext pre-training and downstream domain knowledge to select the most instructive samples.
DOKT consists of a \emph{traceback diversity indicator} for exploring the data distributions in high- and low-level feature spaces and a \emph{domain-based uncertainty estimator} for enforcing perceptual perturbations to measure the degree of data uncertainty.

With respect to the traceback diversity indicator, we pre-train a pretext encoder, which learns the pretext space of the low-level vision task.
Then, we employ intra-class mixing to embed the domain knowledge of annotations into the downstream encoder to alleviate the annotation inadequacy problem encountered when constructing the downstream high-level data representation space.
With the high-quality encoders, we build a traceback module to transfer the adaptive neighbors of unlabeled data from the low-level pretext space to the high-level downstream space, thereby obtaining the neighbor relations of the high- and low-level distributions.
Traceback diversity overcomes the lack of domain knowledge in the pre-training task for comprehensive diversity estimation purposes.
For the domain-based uncertainty estimator, we design a perceptual perturbation method to introduce the pretext neighbors as perturbations and force them to construct domain-mixing augmentations.
With this, we build an attached uncertainty model to determine the uncertainty level using a ranking loss, and the domain uncertainty can be used to jointly estimate the domain divergence of perturbed probabilities.
Benefiting from the traceback diversity and domain uncertainty, DOKT selects the most diverse unlabeled samples and instructive samples of both pretext and downstream domain knowledge.

To demonstrate the effectiveness of DOKT, we conduct comprehensive experiments on ten public datasets. DOKT outperforms state-of-the-art AL methods, showing it is a good fit to facilitate AL for downstream tasks. 
Further, the ablation study also verifies the contribution of the traceback diversity indicator and domain-based uncertainty estimator. 
Besides, since DOKT leverages domain knowledge of annotation for selection, it is adaptive to various downstream applications, such as semantic segmentation and multi-modal tasks.
Thus, we extend DOKT to semantic segmentation and image captioning, in which DOKT still achieves superior performance.


The main contributions of this paper are summarized as:
\begin{enumerate}[label=(\roman*), itemindent=1em]
	\item We propose a downstream-pretext domain knowledge traceback (DOKT) AL method. It unifies pre-training guidance and domain knowledge to explore data distribution for selecting the most instructive samples.
	\item We design a traceback diversity indicator to trace the data relationships in both the low-level and high-level spaces to evaluate the diversity of unlabeled data with comprehensive knowledge.
	\item We design a domain-based uncertainty estimator that introduces domain disturbance by uncertainty mixing to jointly estimate the divergence for selecting the most uncertain samples near the decision boundary.
	\item Extensive experiments demonstrate that DOKT outperforms the SOTA methods and generalizes on multi-modal tasks. The visualization also verifies the effectiveness of DOKT to deeply analyze the AL mechanism.
	
\end{enumerate}


\section{Related work}
\label{sec:2}
Due to the growing demand for large-scale datasets and abundant data sources, most contemporary AL works focus on pool-based AL, which can be categorized as distribution-based and uncertainty-based approaches. 

Distribution-based approaches choose data points that can increase the diversity of labeled data.
These methods usually measure diversity based on extracted representations.
Some methods~\cite{yang2015multi, hasan2015context,chang2018active,sener2017active} select diverse samples by mapping the distribution distance to the informativeness of a data point.
Moreover, some approaches estimate the diversity of a distribution based on the gradients~\cite{settles2008multiple, agarwal2020contextual}, errors~\cite{roy2001toward,beluch2018power}, and output changes~\cite{kading2016active} of trained models or visual information~\cite{yan2016image,WangLMMGZ20}.
CCAL~\cite{du2021contrastive} determines the selected data by evaluating class
distribution mismatches, and BADGE~\cite{Ash2020Deep} selects samples in a gradient space.
However, these methods heavily depend on the supervision of labeled data, while inadequate labeled data in the early sampling stages affect the representation quality.

Uncertainty-based approaches estimate data uncertainty and select the most uncertain samples.
The traditional methods estimate uncertainty by calculating the confidence of classifier errors~\cite{chatzilari2016salic}, Gaussian processes~\cite{kapoor2007active, zhao2021uncertainty} or discrepancies~\cite{huang2021semi,ben2022tackling}.
Several studies~\cite{wu2017weak, farquhar2021statistical} have analyzed the experimental design criteria and biases in statistics.
These traditional methods perform well in some specific tasks but are not effective in cases with deep networks or large-scale datasets.
Some recent works have designed ensemble models~\cite{beluch2018power} or introduced Monte Carlo dropout~\cite{yang2017suggestive}, but these methods are computationally inefficient and unstable for uncertainty estimation tasks.
Yoo \emph{et al.}~\cite{yoo2019active} designed a prediction loss learning model and inspired some related works~\cite{kim2021task, xie2022towards}.
Their method consists of a target module and a loss prediction module for predicting the target loss.
However, the above methods utilize only limited annotation information, and their loss prediction accuracy heavily depends on the generalizability of the target model.
Core-GCN~\cite{caramalau2021sequential} utilizes graph embeddings to calculate confidence scores and adapts other AL techniques; however, this approach is easily affected by a disorganized unlabeled pool.

Recently, some AL methods~\cite{sinha2019variational,Zhang0YWZH20, WangLMMGZ20, choi2021vab, kim2021task} train the encoder in an unsupervised manner and distinguish informative samples in an adversarial manner.  
They map the unlabeled samples to a labeled/unlabeled state label while the binary label is not equivalent to informativeness and cannot reflect the relative importance.
Moreover, the representation by VAE trained on unlabeled pool is also unreliable for AL sampling.
Some AL methods~\cite{bhatnagar2020pal,parvaneh2022active, yi2022pt4al} utilize the Vision Transformer (ViT)~\cite{dosovitskiy2020image} or supervised pretext tasks. 
ALFA~\cite{parvaneh2022active} trains the ViT by labeled samples and mixes the representation of unlabeled samples to estimate the inconsistency. 
However, the labeled samples in AL iterations are usually insufficient to support the training of the Transformer network.
Besides, the random mixing of two different images may also result in heavy inconsistency and lead to inefficient sampling.
PAL~\cite{bhatnagar2020pal} and PT4AL~\cite{yi2022pt4al} train an encoder with self-supervised pretext tasks and select samples based on the pretext loss, while the pretext tasks do not involve domain knowledge of manual annotation.
Thus the pretext loss is not a reliable uncertainty criterion.



\section{Method}
\label{sec:3}

\subsection{Overview}
\label{sec:31}
In this section, we formally define the active learning (AL) scenario and provide the notations.
The key task of an AL algorithm is to select the most informative samples for efficient annotation.
We assume that AL serves to annotate samples for training a target model $\Theta$ for a shared target task.
In the initial stage, a large unlabeled data pool is given, and we annotate a few samples to form an initialized labeled pool.
Let us denote the initial unlabeled pool by $D_U$ and the initial labeled pool by $D_L$. $(x_U)$ denotes a data point in the unlabeled pool, and $(x_L, y_L)$ denotes a data point and its annotation in the labeled pool.
In each iteration, the AL method selects $\mathcal{M}$ samples and obtains their annotations via an oracle.
This iterative process repeats until the model performance satisfies user's requirements or the label budget $\mathcal{B}$ runs out.

\begin{figure*}
	\begin{center}
		\includegraphics[width=0.85\linewidth,trim=10 35 220 10]{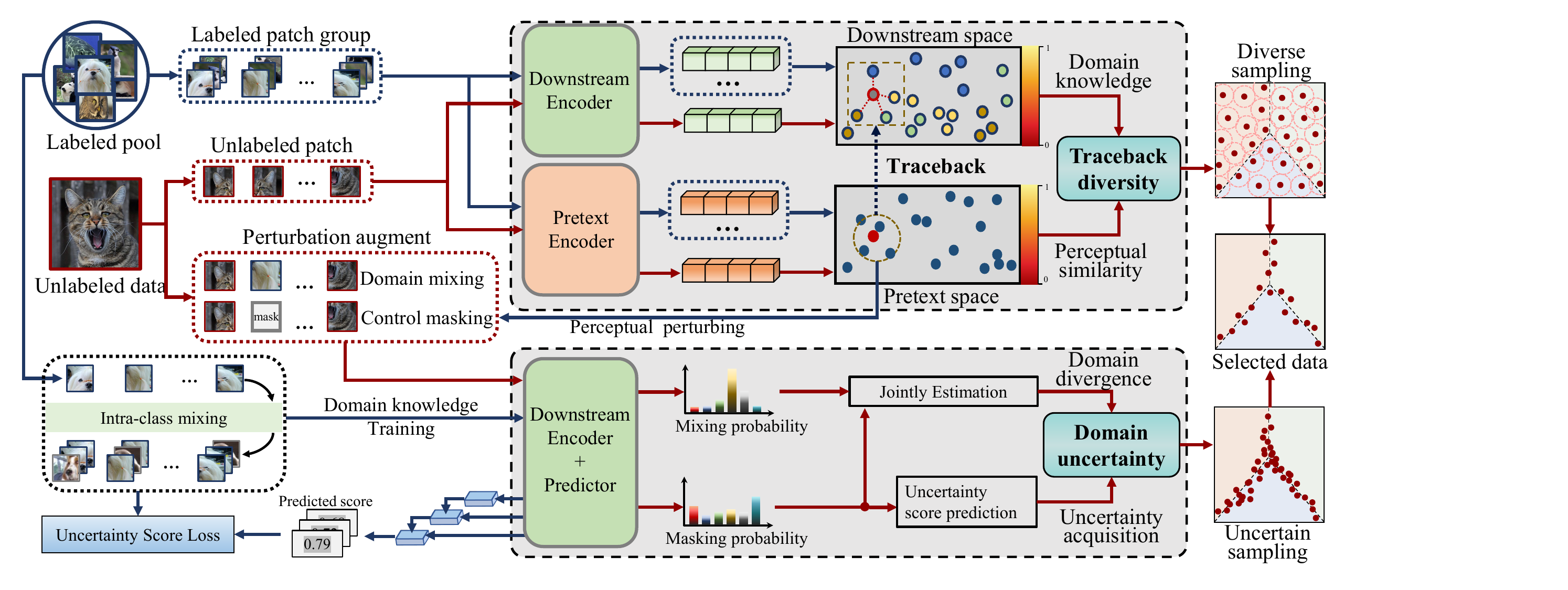}
	\end{center}
	\caption{The network architecture of proposed DOKT.   It consists of a traceback diversity indicator and a domain uncertainty estimator. The indicator traces similar unlabeled samples in two spaces for traceback diversity.  The estimator learns to predict the uncertainty score based on domain mixing samples and apply perceptual perturbing to jointly estimate the divergences of augmented samples for domain uncertainty. Finally, DOKT selects the most diverse and instructive samples based on the two modules. The red lines denote pipelines for unlabeled data flow and the blue lines for labeled data flow.}
	\label{img2}
\end{figure*}

Fig.~\ref{img2} shows the downstream-pretext domain knowledge traceback (DOKT) model.
In order to select most diverse and instructive samples, it traces data interaction from the low-level feature of pre-training guidance to the high-level feature of downstream domain knowledge.
The DOKT consists of a traceback diversity indicator (Section~\ref{sec:32}) and a domain-based uncertainty estimator (Section~\ref{sec:33}). 
The former explores data relationships in two feature spaces to select the most diverse samples, and the latter estimates domain uncertainty based on probability divergence and uncertainty acquisition to select samples near the decision boundary.
A sampling strategy based on the indicator and estimator is introduced in Section~\ref{sec:34}.

As DOKT learns the data relations with pretext and downstream domain knowledge, DOKT generalizes on different downstream tasks with more data modalities~\cite{liu2022entity,deng2021syntax}.
To verify this, we extend DOKT to the semantic segmentation and multi-modal image captioning AL scenarios, which are more challenging and require labeled datasets with higher quality to learn a competitive target model.

\subsection{Traceback diversity indicator}
\label{sec:32}
To comprehensively estimate the diversity of unlabeled samples, we introduce a pre-trained model as a traceback diversity indicator for better learning representations.
For the indicator, we set a pretext encoder $\Theta_p$, which is a vision transformer (ViT)~\cite{dosovitskiy2020image} network parameterized by the weight of the MAE of the pre-training model~\cite{he2022masked}.
Following the transformer, the input of the pretext encoder is a sequence of image patches $\{ x_{i}^{p} \}_{i=1}^{N}$.
Since the pre-training model learns from self-supervised pretext tasks without annotations, it involves only low-level visual perception and lacks domain knowledge concerning the manual annotations.
Thus, directly using the pre-training representations in the AL sampling process would result in the selection of samples that are diverse only in the pretext space and not instructive for learning downstream annotations.
Thus, since the rudimentary process of using the pre-trained model is inefficient, we train a downstream encoder $\Theta_d$ based on $\Theta_p$ and embed domain knowledge in it.

Intuitively, to embed the encoder $\Theta_d$ with domain knowledge of downstream annotation, the model should be trained by the labeled samples in downstream tasks.
However, AL sampling is an iterative process and at its early stages the labeled samples are inadequate to train a robust encoder. 
Inspired by the mixup~\cite{zhang2017mixup}, we design the intra-class mixing to mix up the patches of different labeled samples.
Due to annotation inadequacy, random mixing is hard to learn and sometimes there is no valid object in the mixed image in the result of the random process.
In view of this, the intra-class mixing method only mixes samples that are in the same category and locate closely in the pretext feature space.
With its help, we augment the labeled samples and make the encoder $\Theta_d$ easier to fit.
The training loss of $\Theta_d$ is formulated as follows,
\begin{equation}\label{eq1}
	\begin{aligned}
		\mathcal{L}_d &= {\rm Cross}\text{-}{\rm entropy}(f(Mixing(x_{L}, x_{L}'); \Theta_d), y_L),  \\
		&\text{where} ~ y_L = y_L' ~ \text{and} ~ {\rm Cosine}(x_{L}, x_{L}'; \Theta_p) < \tau
	\end{aligned}
\end{equation}
where $f(;\Theta)$ is a predictor with encoder $\Theta$, $(x_{L}, y_{L})$ is labeled data, and $\tau$ is the median cosine of sample pairs in a same category.

With the pretext encoder $\Theta_p$ and downstream encoder $\Theta_d$, we obtain the low-level representation $r^p$ and high-level representation $r^d$ of the labeled and unlabeled samples:
\begin{equation}\label{eq2}
	\begin{aligned}
		&r^p_L = \Theta_p (x_L) , ~ r^p_U = \Theta_p (x_U) \\
		&r^d_L = \Theta_d (x_L), ~ r^d_U = \Theta_d (x_U) 
	\end{aligned}
\end{equation}
As the pretext encoder is pre-trained by low-level vision tasks and the downstream encoder is fine-tuned by labeled data, $r^p$ learns low-level perceptual information, and $r^d$ learns high-level information from downstream domain knowledge.

Then, the indicator constructs the pretext space with the low-level representation $r^p_L$ and the downstream space with the high-level representation $r^d_L$.
In the two spaces, we can analyze the relationship of unlabeled samples and calculate the diversity.
Intuitively, we can separately calculate the scores of data diversity in two spaces and sum them up as a synthetic diversity for unlabeled samples. 
However, this synthetic manner is not available for AL sampling.
On one hand, the data distance in the high-level space cannot precisely denote the label similarity because it is not the predicted probability.
On the other hand, diversity is a non-additive quantity, so selection based on the sum of two diversity scores is not effective.

Hence, we design the traceback method to naturally quantify the diversity scores of the two spaces.
For an unlabeled sample $x_U$, we select the $\iota$ closest labeled samples $\{x_i\}_\iota$ in the pretext space as its neighbors.
Then, we trace the labeled neighbors to the downstream space to query their annotations $\{y_i\}_\iota$ and obtain $ \mathcal{X}_{pre}=\{x_i, y_i\}_\iota $, where $y$ is a one-hot label vector whose label is denoted as 1 and whose other elements are 0.
$\iota$ denotes the number of closest samples selected as neighbors.
It is an adaptive parameter, and we design a dynamic approach to optimize its value; through this parameter the neighbors can reflect the diversity of $x_U$ in two spaces.
To optimize $\iota$, the indicator scales the cosine similarity range to [0,1] and averages the cosine similarities of $x_U$ and $ \mathcal{X}_{pre} $ as
\begin{equation}\label{eq3}
	{\rm mcosine}(x_U) = \sum_{(x_i, y_i)\in \mathcal{X}_{pre}} \frac{{\rm Cosine}(x_i, x_U)+1}{2} / \iota
\end{equation}

Then, a traceback domain vector $V_t$ is calculated as
\begin{equation}\label{eq4}
	V_t(x_U) = \sum_{(x_i, y_i)\in \mathcal{X}_{pre}} \frac{{\rm Cosine}(x_i, x_U)+1}{2} \cdot y_i
\end{equation}
where the one-hot label vector $y_i$ of $x_i$ is weighted by the cosine similarity
${\rm Cosine}(x_i, x_U)$; then, the indicator sums the results to obtain the domain distribution of $ \mathcal{X}_{pre}$.

$V_t(x_U)$ represents the domain distribution of the neighbors scaled by the low-level similarity.
With this, the objective function for optimizing $\iota$ is formulated as follows:
\begin{equation}\label{eq5}
	\iota = \underset{\iota \leq L}{\arg \min } \{ {\rm Var(\frac{V_t(x_U)}{\|{V_t(x_U)}\|})} \times {\rm mcosine}(x_U) \}
\end{equation}
where $\iota$ is optimized to minimize the product of the variance and mean similarity, and its maximum value is $L$, which is 1\% of the labeled pool. Eq.~\ref{eq5} is not differentiable or monotonic, so we cannot optimize $\iota$ by directly constraining Eq.~\ref{eq5} or using a binary search.
Thus, we apply enumeration over $\iota$ in $(0,L]$.

Intuitively, this variance quantifies the annotation concentration of the neighbors, and the mean similarity quantifies the diversity in the low-level space.
Thus, the indicator adaptively optimizes $\iota$ to find the set of pretext neighbors that maximizes the diversity of the unlabeled samples $x_U$.
In this way, the indicator calculates the traceback diversity score of $x_U$ as
\begin{equation}\label{eq6}
	S_{trace}(x_U) =  {\rm Var(\frac{V_t(x_U)}{\|{V_t(x_U)}\|})} \times {\rm mcosine}(x_U) 
\end{equation}

\subsection{Domain-based uncertainty estimator}
\label{sec:33}

With the help of its traceback diversity indicator, DOKT traces the data interactions in two spaces and calculates a comprehensive diversity score based on the low- and high-level features.
To improve the AL sampling process, DOKT combines both diversity and uncertainty for selection purposes.
Therefore, we propose a domain-based uncertainty estimator.
Inspired by the learning loss method~\cite{yoo2019active}, we design a multiperception uncertainty learning model $\Theta_U$ to predict the uncertainty of data based on hidden features derived from transformer blocks.
As the CE loss focuses only on the ground-truth element and ignores other probabilities, we design an uncertainty score for $\Theta_U$ to learn. The uncertainty score is formulated as follows:
\begin{equation}\label{eq7}
	S(x_U) = 1 -  \frac{\max (P(x_U)) \times \operatorname{maxVar} P(x_U) }{\operatorname{Var}(P(x_U))} 
\end{equation}
where $P(x) = f(x_U;\Theta_d)$ is the predicted probability vector and $\max (P(x_U))$ is the maximum probability vector. $\operatorname{maxVar}(P(x_U))$ is defined as follows:
\begin{equation}\label{eq71}
	\begin{aligned}
		&\operatorname{maxVar}(P(x_U)) =\\
		&\frac{1}{C}((\frac{1}{C} \!-\! \max (P(x_U)) )^{2} \!+\! (C \!-\! 1)(\frac{1}{C} \!-\! \frac{1 \!-\! \max (P(x_U))}{C \!-\! 1})^2)
	\end{aligned}
\end{equation}
where $C$ is the number of classes.

$\operatorname{maxVar}(P(x_U))$ is the variance for a probability vector whose maximum is the same as that of $P(x_U)$, and the other probabilities are equal to $\frac{1 \!-\! \max (P(x_U))}{C \!-\! 1}$.
It quantifies the highest concentration of a probability vector.
By combining $\operatorname{maxVar}(P(x_U))$ and $\max (P(x_U))$, the result is negatively correlated with the maximum probability and concentration of the probability vector, providing a better uncertainty acquisition.
Based on $S(x_U)$, we formulate a ranking loss to update $\Theta_U$:
\begin{equation}\label{eq8}
	\mathcal{L}_r(\widehat{s_1},\widehat{s_2}) =  \left \{
	\begin{aligned}
		0 ~ ~ ~ ~ ~ ~ ~ ~ ~ ~ ~ ~ ~ & ~ \text{if } ~ | \widehat{s_1} - \widehat{s_2} | > l_{margin}   \\
		(\widehat{s_1} - \widehat{s_2})(s_1 - s_2)& ~ \text{ otherwise}
	\end{aligned}
	\right.
\end{equation}
where $s_1 = S(x_1)$ and $s_2 = S(x_2)$ are the scores calculated using Eq.~\ref{eq7}; $\widehat{s_1}$ and $\widehat{s_2}$ are the scores predicted by $\Theta_U$ using the hidden features in $\Theta_d$; and $l_{margin} $ is a margin.
As $S(x_U)$ is limited to {[0,1)}, updating with the ranking loss is easy to converge.
Assume that $s_1 > s_2$, minimizing $L_r$  enlarges $\widehat{s_1} - \widehat{s_2}$ unless their difference is larger than the margin.
Compared with MSE, this loss using ranking relation is easy to converge.
Meanwhile, the ranking loss also constrains the concrete difference $s_1 - s_2$ to ensure that the range of output in $\Theta_U$ is similar to the uncertainty score.
In this way, $L_r$ utilizes the ranking information and the ground-truth difference $s_1 - s_2$ to optimize uncertainty prediction.
By training the uncertainty model via Eq.~\ref{eq8}, the estimator can extract hidden features through $\Theta_d$ and precisely quantify uncertainty with $\Theta_U$:
\begin{equation}\label{eq9}
	S(x_U) = f(x_U; \Theta_U, \Theta_d)
\end{equation}

Further, we enforce domain mixing on unlabeled samples to apply perceptual perturbation.
In ALFA-Mix~\cite{parvaneh2022active}, it mixed transformer patches by randomly mixing representations via interpolation to calculate the inconsistency as the uncertainty.
However, the mixing process for representation confuses the embedded information and makes it difficult to learn, and the random mixing strategy may mix two highly different images so that the mixing results exhibit strong inconsistency while remaining uncertain for the target model.

In view of this, we design a perturbation augmentation process to mix unlabeled samples with similar labeled data to form an applicable mixing strategy.
For an unlabeled sample $x_U$, the estimator mixes its patches with its neighbors in the pretext space to form domain mixing samples.
Then, the estimator duplicates the domain mixing samples and replaces the mixed patches with masked patches.
In this way, the model obtains the domain mixing samples $Mixing(x_U, \mathcal{X}_{pre})$ and the corresponding masking samples $Masking(x_U, \mathcal{X}_{pre})$.
The downstream encoder is fine-tuned with a labeled pool, as shown in Eq.~\ref{eq1}, to predict the probability vectors $v_{mix} = f(Mixing(x_U, \mathcal{X}_{pre}); \Theta_U)$ and $v_{mask} = f(Masking(x_U, \mathcal{X}_{pre}); \Theta_U)$ for the two samples.
Based on these vectors, the estimator calculates the divergence for these two probabilities as follows:
\begin{equation}\label{eq10}
	D(x_U) = D_{KL}(x_U) =\sum_{i} v_{mix}(i) \log (\frac{v_{mix}(i)}{v_{mask}(i)})
\end{equation}
where KL divergence~\cite{1057082} and $D(x_U)$ are used to quantify the data uncertainty when the data are mixed with similar labeled samples. A smaller $D(x_U)$ value denotes a more certain prediction, and
a larger $D(x_U)$ value means that the perceptual perturbation process implemented via similar samples makes the prediction unsteady and more uncertain.

By summing up the uncertainty acquisition as Eq.~\ref{eq9} and domain divergence as Eq.~\ref{eq10}, the estimator can estimate the domain uncertainty of unlabeled samples to maximize the performance gain in each sampling iteration:
\begin{equation}\label{eq11}
	D_{domain}(x_U) = S(x_U) + D(x_U)
\end{equation}

Using Eq.~\ref{eq11}, DOKT analyzes the data uncertainty with perceptual perturbing and domain divergence.
By combining the domain uncertainty $D_{domain}(x_U)$ and traceback diversity in two spaces, DOKT can select instructive samples near the decision boundary to boost modal training and avoid overlapping of selected samples.

\subsection{Sampling strategy in active learning}
\label{sec:34}

DOKT consists of two steps in each sampling iteration: model training and sample selection.
In the training stage, the indicator trains the downstream encoder using intraclass mixing and augmentation to embed domain knowledge.
Along with this training process, the estimator trains the attached uncertainty model using Eq.~\ref{eq7} by mixing labeled samples, and the model learns to predict an uncertainty score based on the hidden features in transformer blocks.

In the selection stage, the indicator quantifies the traceback diversity in the pretext and downstream spaces.
The estimator learns to predict uncertainty scores using the ranking loss and performs perceptual perturbation to produce augmentations.
Then, the divergence of probabilities is calculated to estimate the prediction stability.
By combining the uncertainty score and domain divergence, the domain uncertainty can be estimated.
With these two modules, DOKT first filters out the $2\mathcal{M}$ most diverse samples as candidates based on the traceback diversity and subsequently selects the $\mathcal{M}$ most uncertain samples among the candidates based on the domain uncertainty.

\section{Experiment}
\label{sec:4}

\begin{figure*}
	\centering
	\subfloat[{\footnotesize  CIFAR-10}]{\includegraphics[width=0.22\linewidth]{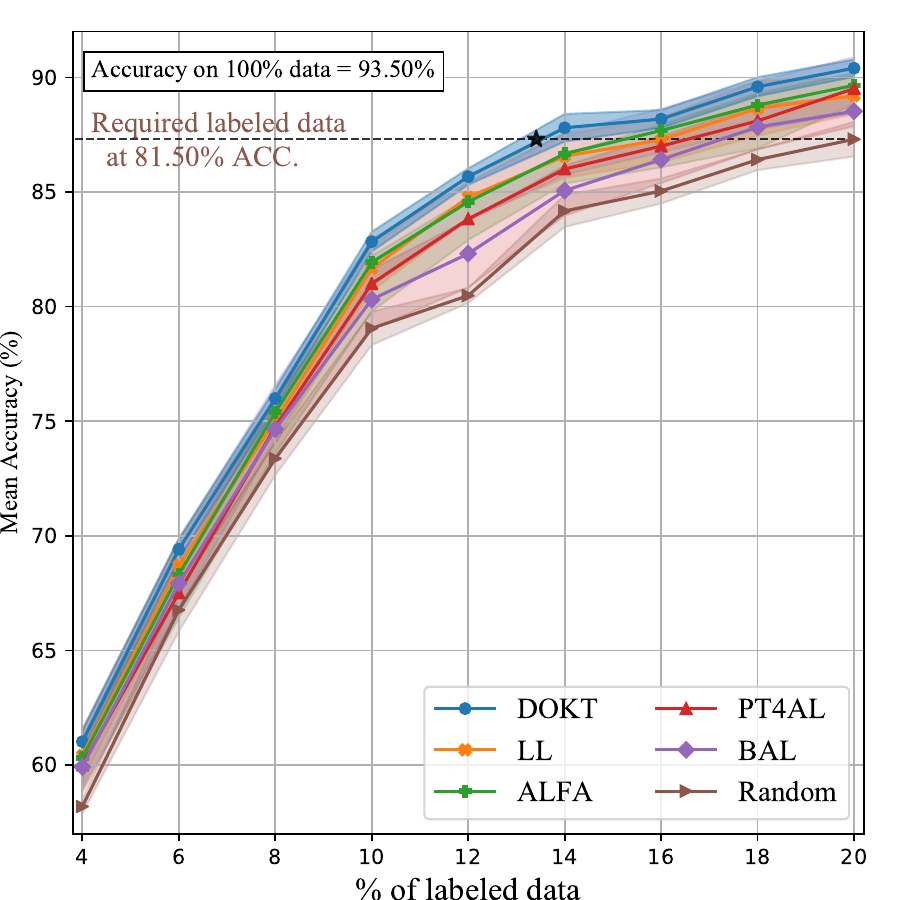}} 
 \qquad
	\subfloat[{\footnotesize  CIFAR-100}]{\includegraphics[width=0.22\linewidth]{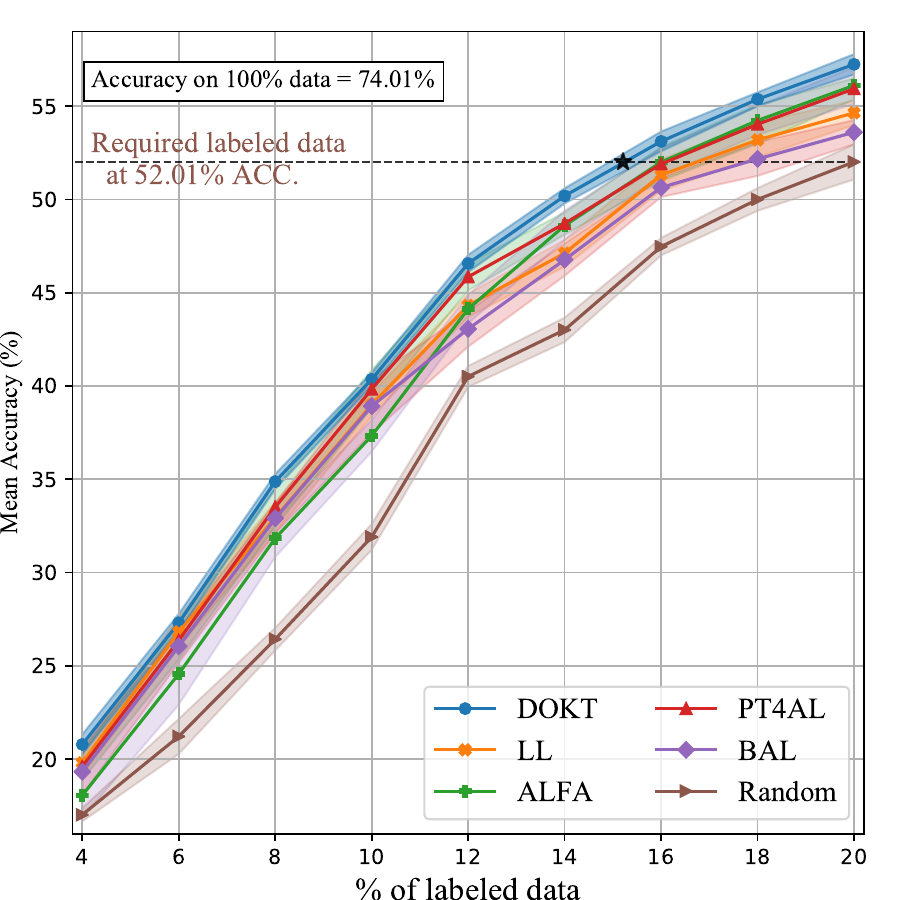}} 
  \qquad
	\subfloat[{\footnotesize  SVHN}]{\includegraphics[width=0.22\linewidth]{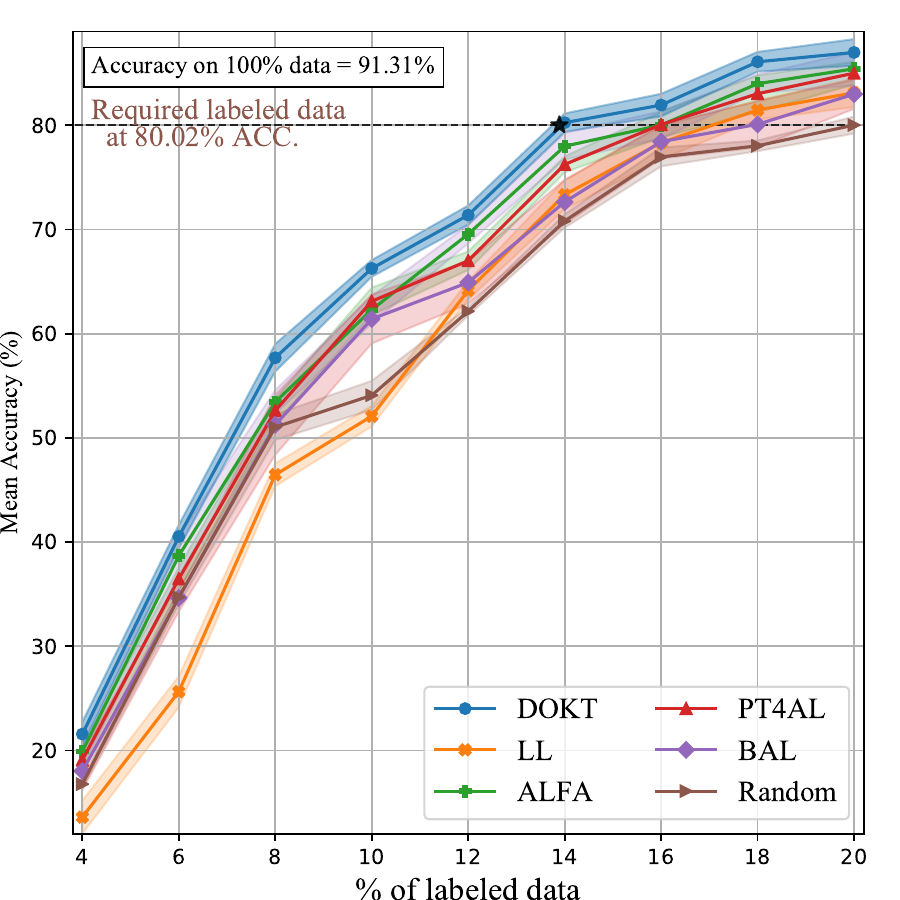}} 
  \qquad
	\subfloat[{\footnotesize  Aircraft}]{\includegraphics[width=0.22\linewidth]{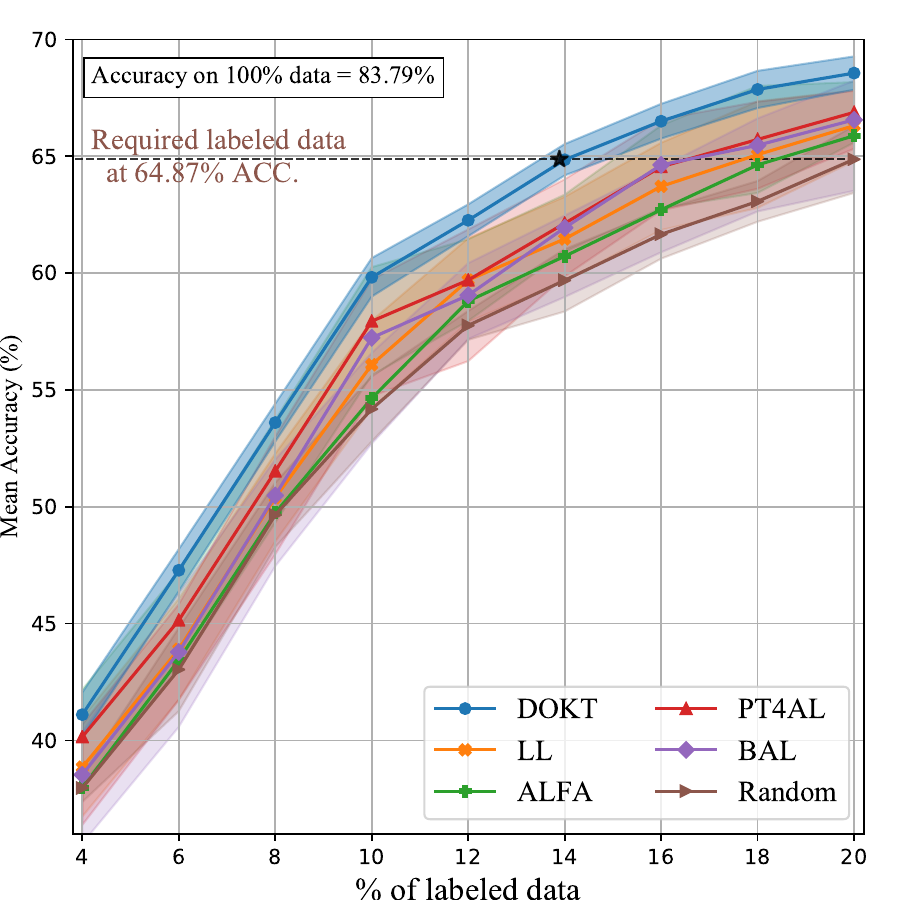}} 
	
	\caption{Results of different AL methods on the image classification datasets. (a) and (b) are the results on CIFAR-10 and CIFAR-100. (c) is the result on SVHN. (d) is the result on the fine-grained dataset Aircraft. The dotted line shows the required labeled data to achieve a competitive performance and the required annotation of AL methods is detailed in Table~\ref{tabnum}.}
	\label{f1}
\end{figure*}

\subsection{Experiment Settings}
\label{sec:41}

In this section, we compare the performance of our method with those of the latest AL methods on ten datasets.
Following other AL works~\cite{yoo2019active, sinha2019variational}, we conduct our experiments on two common tasks: image classification and semantic segmentation.
For classification purposes, we also test on fine-grained data to evaluate the performance achieved for images with subtle downstream categorization differences.
As DOKT is also effective in cross-modal AL tasks, we explore the AL sampling for image captioning.
Moreover, we conduct experiments on long-tailed datasets to verify the generalizability of the proposed method to problems with categorical imbalances.

For rigorous comparison, all the AL methods begin with the same random initial labeled pool and they are evaluated by the same target model. 
Since the performance achieved with the initial labeled data is fixed for all the AL methods, we only provide the performance starting from the first iteration. 

%

\subsection{Experimental Results on Image Classification} 
~\label{4.1}
\noindent \textbf{Dataset.} 
We test on CIFAR-10, CIFAR-100~\cite{krizhevsky2009learning}, SVHN~\cite{netzer2011reading} and the fine-grained Aircraft dataset~\cite{maji2013fine}.
CIFAR-10 and CIFAR-100 have 60,000 images with sizes of 32$\times$32$\times$3, including 50,000 training images and 10,000 testing images. The CIFAR-10 has 10 categories and 6,000 images per class, while the CIFAR-100 has 100 classes containing 600 images each.
SVHN is a real-world image dataset with 73,257 digits for training and 26,032 digits for testing.
For SVHN, we test on the original images with multiple digits.
Aircraft is a fine-grained classification dataset that contains 100 images for each of the included 100 aircraft variants.
We split Aircraft into 6,667 training samples and 3,333 test samples.

\noindent \textbf{Compared methods.} 
We compare DOKT with several recently developed state-of-the-art approaches, including LL~\cite{yoo2019active}, PT4AL~\cite{yi2022pt4al}, ALFA~\cite{parvaneh2022active}, and BAL~\cite{zhang2023scalable}.
We introduce the random sampling method as a baseline.

\noindent \textbf{Performance measurement.}
Following the settings of other AL methods, in the beginning, we randomly select 2\% of the data as the initial labeled pool.
We rigorously report the mean performance achieved across 5 trials with different random initializations.
In each iteration, 2\% of the samples are selected to be labeled, and the label budget is 20\% of the dataset.
The target model is an 18-layer residual network (ResNet-18)\cite{7780459}.

\subsubsection{Performance on CIFAR-10}
Fig.~\ref{f1}~(a) shows the AL performances achieved on CIFAR-10.
First, DOKT outperforms the other state-of-the-art methods under all ratios.
The consistent margin indicates that DOKT with the traceback module can efficiently leverage its pre-trained model and select more informative samples than can the other AL methods.
Second, the accuracy of DOKT reaches 90.35\% when using 20\% of the data, while the highest accuracy achieved on fully labeled data is 93.5\%, which is only 3.15\% better than that of DOKT with an 80\% annotation cutoff.
This competitive performance demonstrates that DOKT can significantly improve the quality of labeled datasets for robustly training models.

\subsubsection{Performance on CIFAR-100}
As CIFAR-100 has more categories and fewer images per category, it is more challenging than CIFAR-10 is.
Fig.~\ref{f1}~(b) shows the performances achieved on CIFAR-100.
As observed with CIFAR-10, DOKT still performs better than the other AL methods.
In addition, several different conclusions can be drawn.
First, we observe that the uncertainty-based LL method~\cite{yoo2019active} does not perform as competitively on CIFAR-100 as it does on CIFAR-10.
This indicates that insufficient domain knowledge restrains the backbone model and leads to an inefficient sampling process.
Moreover, larger margins are produced between DOKT and the other AL methods because domain mixing augments the limited samples for general model training.
Second, PT4AL~\cite{yi2022pt4al} surpasses ALFA~\cite{parvaneh2022active} in the early iterations since ALFA trains a ViT with labeled samples and because the inadequacy of the annotations in the early stages invalidate its estimation results.
This observation demonstrates that DOKT learns correct data interactions with limited annotations by tracing back two steps and generalizing to datasets with more categories.

\begin{figure*}
	\centering
	\subfloat[{\footnotesize  CIFAR-10-LT}]{\includegraphics[width=0.22\linewidth]{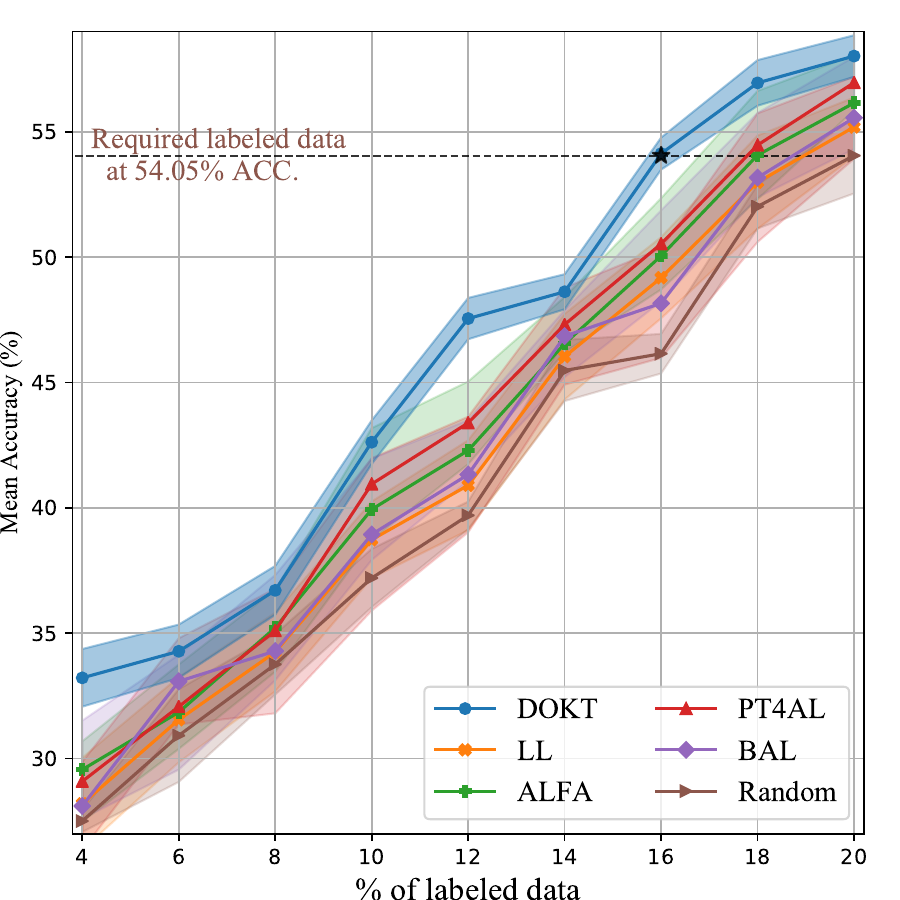}} 
 \qquad
	\subfloat[{\footnotesize  CIFAR-100-LT}]{\includegraphics[width=0.22\linewidth]{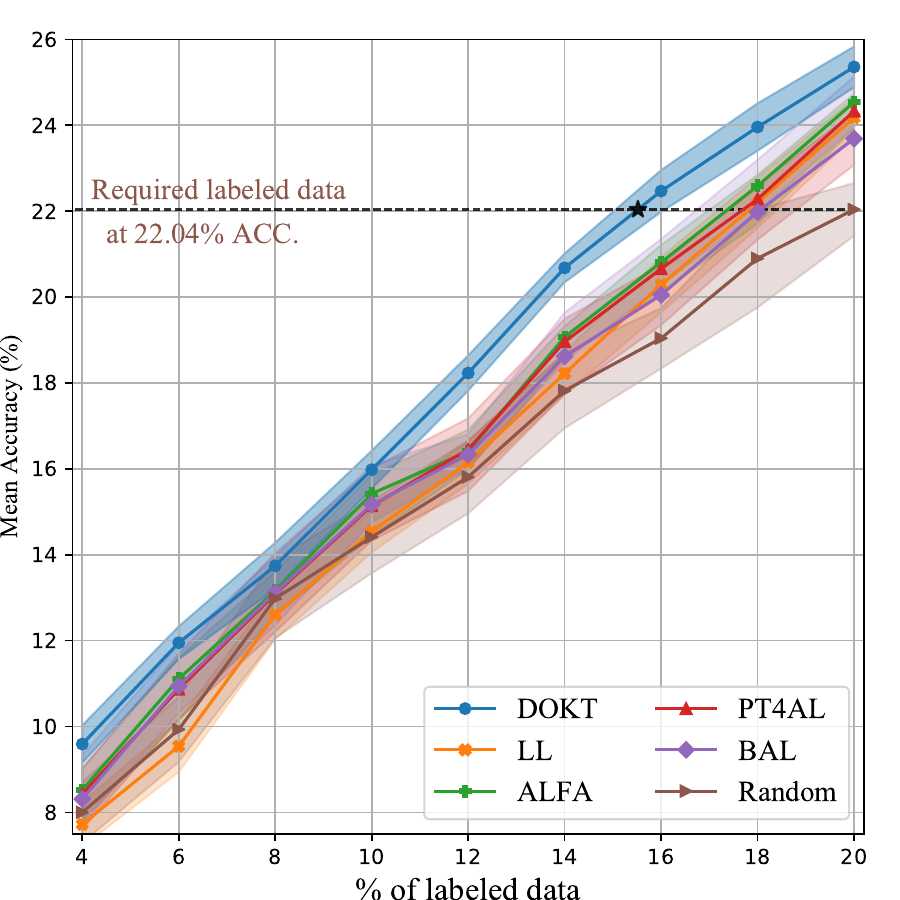}} 
 \qquad
	\subfloat[{\footnotesize  ImageNet}]{\includegraphics[width=0.22\linewidth]{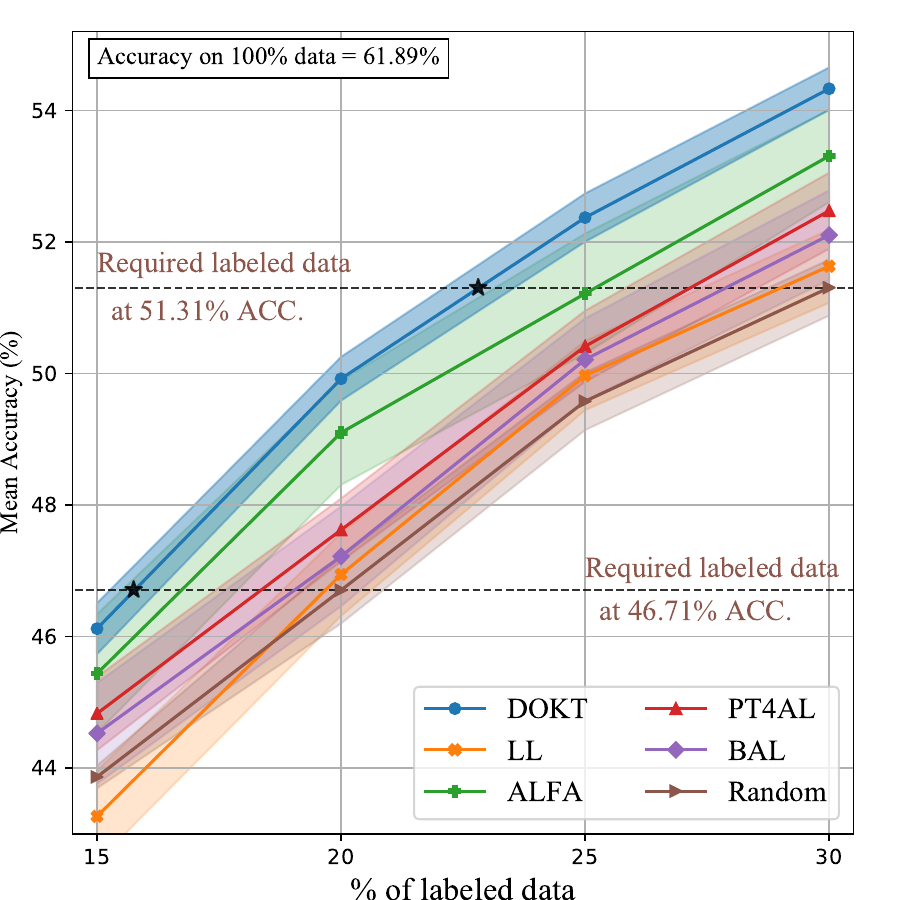}} 
 \qquad
	\subfloat[{\footnotesize  iNaturalist}]{\includegraphics[width=0.22\linewidth]{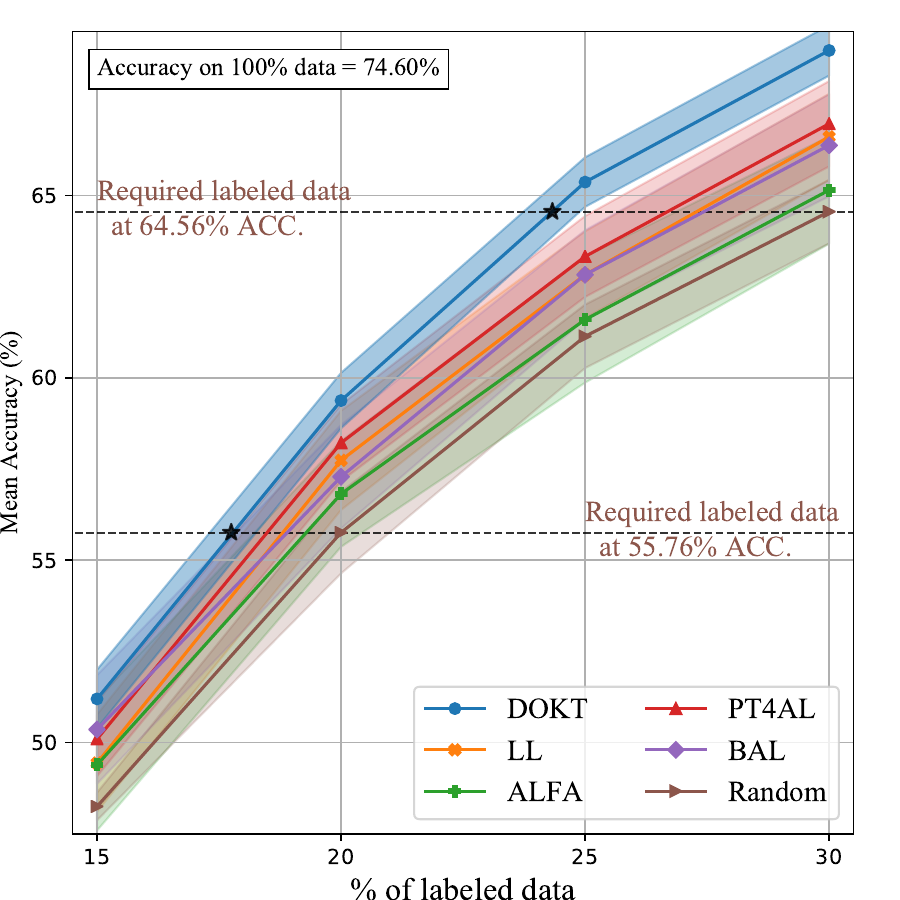}} 
	
	\caption{Results of different AL methods. (a) and (b) are the results on long-tailed CIFAR-10 and CIFAR-100. (c)  is the result on ImageNet. (d) is the result on iNaturalist.}
	\label{f2}
\end{figure*}

\begin{table*}
	\centering
	\caption{The required number of labeled samples to achieve the target performance using 20\% selected samples for different AL methods.}
	\begin{tabular}{c|c|c|c|c|c|c|c|c}
		\hline
		Dataset & CIFAR-10 & CIFAR-100 & SVHN & Aircraft &  CIFAR-10-LT & CIFAR-100-LT  & ImageNet & iNaturalist \\
        (Target ACC) & (81.50\%) & (52.01\%) & (80.02\%) & (64.87\%) & (54.05\%)\ & (22.04\%)  & (46.71\%) & (55.76\%) \\
		\hline
		Random & 10,000  & 10,000  & 17,305  & 1,333  & 2,482  & 2,179  & 384,350  & 131,253  \\
		LL~\cite{yoo2019active}    & \cellcolor[rgb]{ .988,  .894,  .839}8,049  & \cellcolor[rgb]{ .988,  .894,  .839}9,039  & \cellcolor[rgb]{ .988,  .894,  .839}15,632  & \cellcolor[rgb]{ .988,  .894,  .839}1,197  & \cellcolor[rgb]{ .988,  .894,  .839}2,353  & \cellcolor[rgb]{ .988,  .894,  .839}1,998  & \cellcolor[rgb]{ .988,  .894,  .839}366,413  & \cellcolor[rgb]{ .988,  .894,  .839}120,316  \\
		BAL~\cite{zhang2023scalable}   & \cellcolor[rgb]{ .988,  .894,  .839}8,571  & \cellcolor[rgb]{ .973,  .796,  .678}8,477  & \cellcolor[rgb]{ .988,  .894,  .839}14,891  & \cellcolor[rgb]{ .973,  .796,  .678}1,121  & \cellcolor[rgb]{ .988,  .894,  .839}2,337  & \cellcolor[rgb]{ .988,  .894,  .839}2,012  & \cellcolor[rgb]{ .988,  .894,  .839}358,726  & \cellcolor[rgb]{ .988,  .894,  .839}120,705  \\
		PT4AL~\cite{yi2022pt4al} & \cellcolor[rgb]{ .988,  .894,  .839}8,428  & \cellcolor[rgb]{ .973,  .796,  .678}8,047  & \cellcolor[rgb]{ .973,  .796,  .678}13,930  & \cellcolor[rgb]{ .973,  .796,  .678}1,112  & \cellcolor[rgb]{ .973,  .796,  .678}2,221  & \cellcolor[rgb]{ .973,  .796,  .678}1,945  & \cellcolor[rgb]{ .988,  .894,  .839}345,914  & \cellcolor[rgb]{ .973,  .796,  .678}118,129  \\
		ALFA~\cite{parvaneh2022active}  & \cellcolor[rgb]{ .973,  .796,  .678}7,650  & \cellcolor[rgb]{ .973,  .796,  .678}8,033  & \cellcolor[rgb]{ .973,  .796,  .678}13,945  & \cellcolor[rgb]{ .988,  .894,  .839}1,228  & \cellcolor[rgb]{ .973,  .796,  .678}2,237  & \cellcolor[rgb]{ .973,  .796,  .678}1,892  & \cellcolor[rgb]{ .973,  .796,  .678}326,313  & \cellcolor[rgb]{ .988,  .894,  .839}126,441  \\
		DOKT  & \cellcolor[rgb]{ .957,  .69,  .518}6,552  & \cellcolor[rgb]{ .957,  .69,  .518}7,414  & \cellcolor[rgb]{ .957,  .69,  .518}11,853  & \cellcolor[rgb]{ .957,  .69,  .518}904  & \cellcolor[rgb]{ .957,  .69,  .518}1,977  & \cellcolor[rgb]{ .957,  .69,  .518}1,681  & \cellcolor[rgb]{ .957,  .69,  .518}288,390  & \cellcolor[rgb]{ .957,  .69,  .518}105,003  \\
		\hline
	\end{tabular}%
	\label{tabnum}
\end{table*}%

\subsubsection{Performance on SVHN}
On the SVHN dataset, the target model recognizes multiple digits in an image, and the number of digits per image is unknown, which makes it much more difficult.
Fig.~\ref{f1}~(c) compares different AL methods on a multidigit dataset.
First, we find that DOKT outperforms the state-of-the-art AL methods on SVHN by a significant margin.
This benefits from the use of the traceback module, which quantifies low- and high-level diversity to satisfy complex scenarios.
Second, LL~\cite{yoo2019active} and BAL~\cite{zhang2023scalable} perform worse than does the random method in the early sampling stages.
Since it is difficult to learn the uncertainty of multiple digits based on the predicted probability, the above uncertainty-based methods are unreliable.
Especially in early stages, the limited number of available labeled samples intensifies this problem.
In contrast, DOKT utilizes the traceback to bridge two feature spaces, so that it can comprehensively estimate the data distribution and select diverse samples with a complex data format. 

Further, we analyze the required number of labeled samples to achieve competitive performance.
As shown in Table~\ref{tabnum}, DOKT requires only 13.1\% (6,552), 14.8\% (7,414), and 13.7\% (11,853) of the labeled samples to achieve the same performance when using 20.0\% of the randomly selected samples from CIFAR-10, CIFAR-100, and SVHN, respectively.
Compared with state-of-the-art methods, DOKT reduces the numbers of required annotations by 18.6\% (1,497) on CIFAR-10, 12.5\% (1,063) on CIFAR-100, and 20.4\% (3,038) on SVHN.
These results verify that DOKT can improve the annotation efficiency and quality of datasets to boost the model training.

\subsubsection{Performance on Aircraft}
For the Aircraft dataset, the fine-grained samples in each class differ only in terms of subtle optical details, which is more challenging.
Fig.~\ref{f1}~(d) shows the performances.
We find that DOKT outperforms the state-of-the-art methods and that the gap between DOKT and the other methods becomes larger than those observed on CIFAR datasets.
This improvement stems from the use of the traceback module, which locates important samples near the decision boundary to boost the AL process for fine-grained classification.

In addition, we test on Caltech-101, Caltech-256~\cite{li2004caltech}, CINIC-10~\cite{darlow2018cinic}, Tiny-ImageNet and Mini-ImageNet~\cite{imagenet} in the supplementary material. The results also demonstrate that DOKT achieves the best AL performance and can be generalized to various datasets with the help of the traceback.

\subsection{Experimental Results on Long-tailed Datasets} 
The popular datasets are usually artificially balanced with respect to the instances belonging to each class in the training set.
However, real-world datasets usually have highly imbalanced class distributions with long tails~\cite{ChawlaBHK02}.
To evaluate the performance achieved in this practical scenario, we conduct experiments on long-tailed versions of the CIFAR datasets,  CIFAR-10-LT and CIFAR-100-LT.
To simulate long tails, we reduce the number of training samples per class according to an exponential function $n= \frac{n_{i}}{\mu^{i/C}}  $, where $i$ is the class index, $C$ is the number of classes, $n_i$ is the original number of training images and $\mu=100$ is the imbalance factor.
The experimental settings are the same as those used in Section~\ref{4.1}.

Figs.~\ref{f2} (a) and (b) show the results on CIFAR-10-LT and CIFAR-100-LT.
Due to the label imbalance issue, it is more difficult to learn the target model, especially for categories in the tail.
First, DOKT outperforms the other AL methods at all sampling ratios by large margins.
This demonstrates that DOKT can better select the most informative samples from imbalanced datasets with the help of trackback diversity.
Second, ALFA~\cite{parvaneh2022active} and LL~\cite{yoo2019active} perform less competitively in this scenario.
Since the label imbalance makes model training more difficult, the backbones in ALFA and LL become unreliable, affecting their ability to estimate data inconsistency and the learning loss.
In DOKT, domain mixing can augment the inadequate samples of long-tailed categories, and trackback diversity leverages pre-training for robust distribution learning.
These modules can overcome label imbalances and construct high-quality datasets.
Third, as shown in Table~\ref{tabnum}, DOKT requires only 15.9\% (1,977) and 15.4\% (1,681) of total samples to achieve competitive performance on CIFAR-10-LT and CIFAR-100-LT, respectively.
Compared with other methods, the proposed method requires 15.4\% fewer labeled samples on CIFAR-10-LT and 16.5\% fewer samples on CIFAR-100-LT.
These reductions further verify that DOKT can overcome the bottleneck of AL in cases with imbalanced data.
This approach benefits from the domain knowledge of the downstream and pretext spaces to balance a long-tailed distribution.

\begin{figure*}
	\centering
	\subfloat[{\footnotesize  Cityscapes}]{\includegraphics[width=0.22\linewidth]{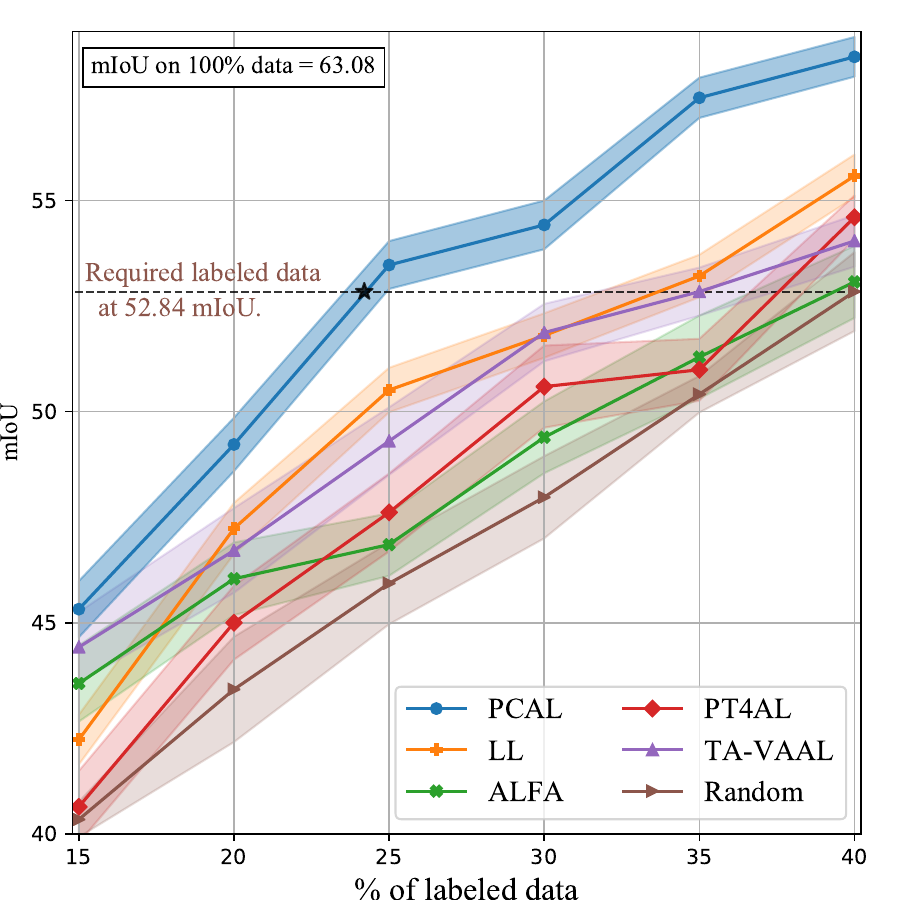}} \qquad
	\subfloat[{\footnotesize  BraTS}]{\includegraphics[width=0.22\linewidth]{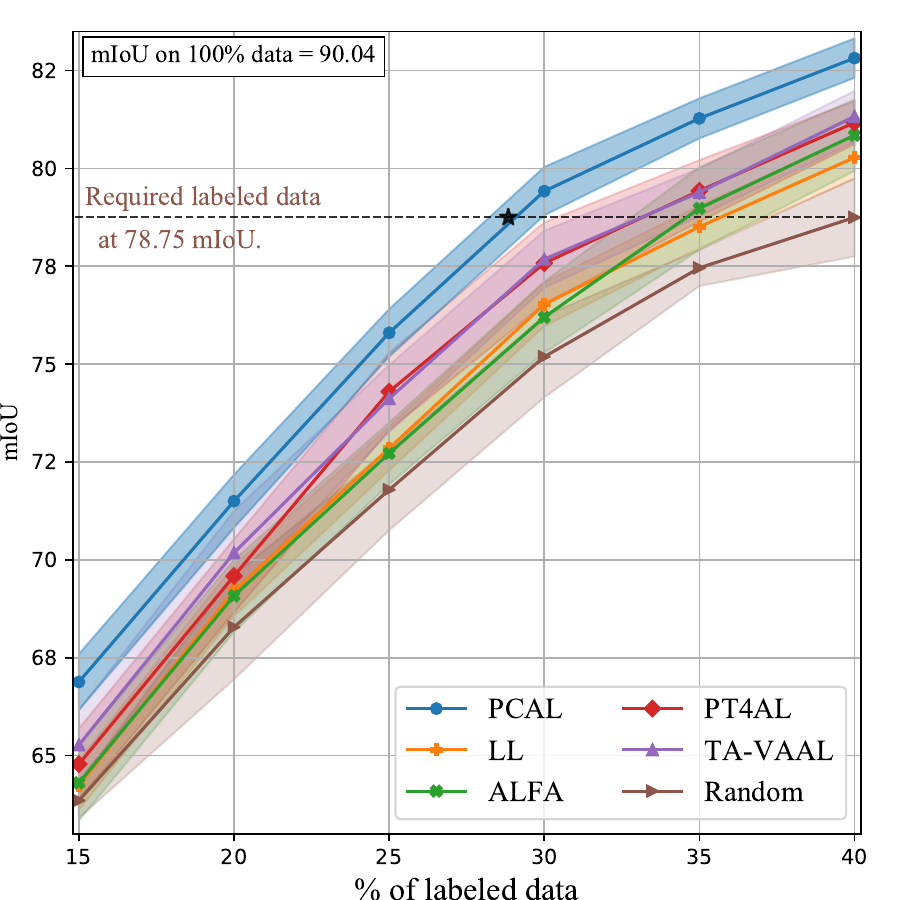}} 
 \qquad
	\subfloat[{\footnotesize  MS-COCO BLEU4}]{\includegraphics[width=0.22\linewidth]{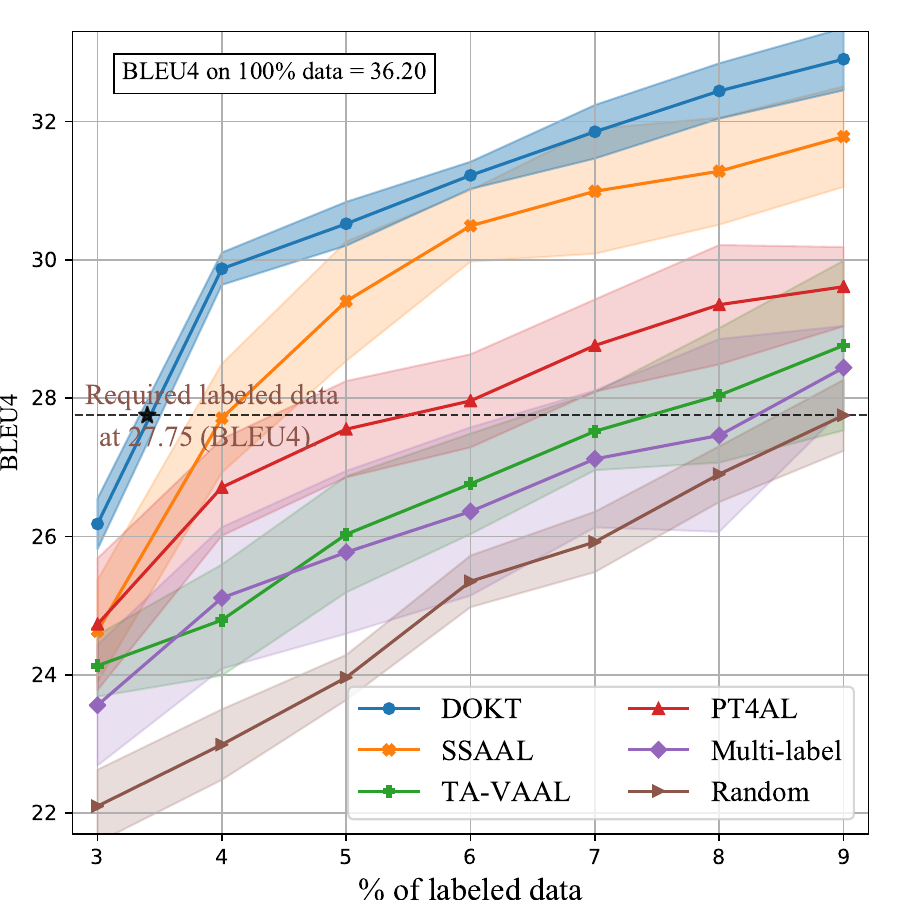}} 
 \qquad
	\subfloat[{\footnotesize  MS-COCO CIDEr}]{\includegraphics[width=0.22\linewidth]{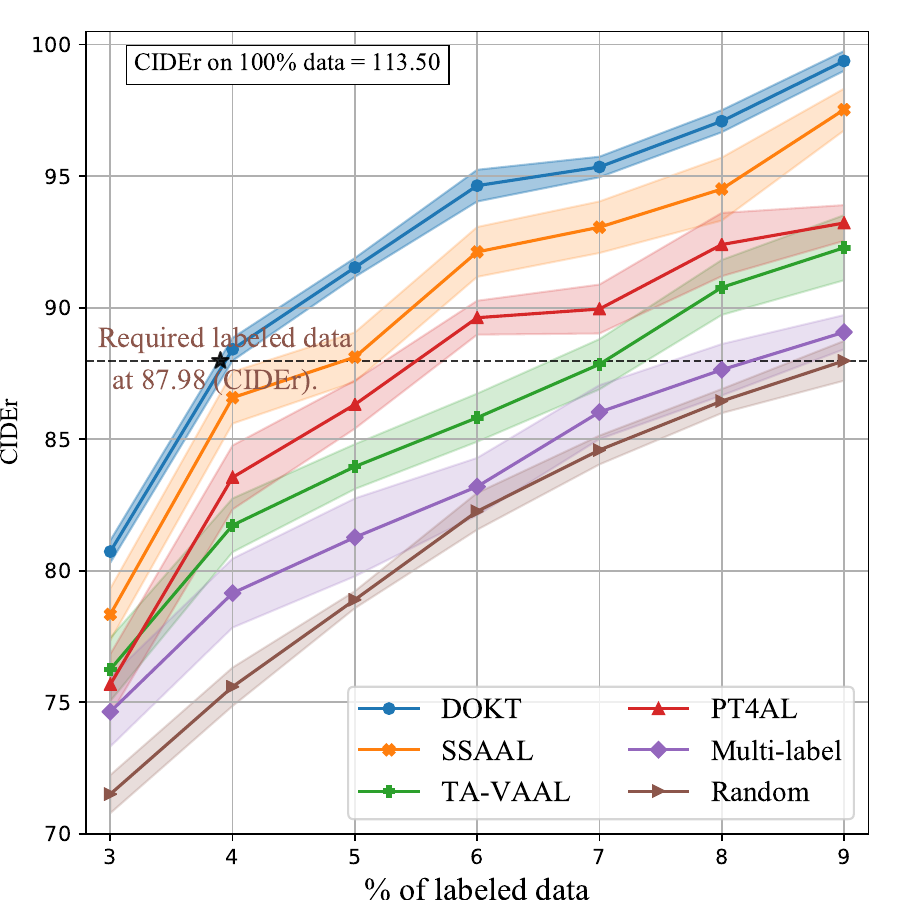}} 
	
	\caption{Results of different AL methods on semantic segmentation and image captioning datasets. (a) and (b) are the results on Cityscapes and BraTS. (c) and (d) are the results on MS-COCO using two metrics BLEU4 and CIDEr.}
	\label{f3}
\end{figure*}

\subsection{Experimental Results on Large-scaled Datasets} 
\noindent \textbf{Dataset.} 
In addition, we demonstrate the scalability of DOKT on two large-scale datasets: ImageNet~\cite{imagenet} and iNaturalist 2018~\cite{van2018inaturalist}.
ImageNet consists of more than 1.2 M images belonging to 1000 classes.
iNaturalist is a large-scale real-world dataset for species classification that contains 437,513 images from 8,142 categories.
Furthermore, iNaturalist is an extremely long-tailed dataset with an imbalanced label distribution.

\noindent \textbf{Compared methods.}
The compared AL methods are the same as those in Section~\ref{4.1}.

\noindent \textbf{Performance measurement.}
Following the settings of previously AL methods, we randomly select 10\% of the available data as the initial labeled pool.
We rigorously report the mean performance achieved across 5 trials with different random initializations.
In each iteration, 5\% of the samples are selected to be labeled, and the label budget is 30\%.

\subsubsection{Performance on ImageNet}
The experimental results are shown in Fig.~\ref{f2} (c).
First, DOKT consistently outperforms all the baselines, with average margins of 3.42\% over random sampling and 1.17\% over the most competitive AL method (PT4AL)~\cite{yi2022pt4al}.
These margins are larger than above observation.
This finding verifies the generalizability of DOKT to different data scales, which benefits from the domain uncertainty measurement process that locates instructive samples near the decision boundary.
Second, as shown in Table~\ref{tabnum}, we find that DOKT achieves the same performance attained when using 30\% of the random data with 22.5\% (288,390) of the samples, which are 25.0\% (95,960) and 16.6\% (57,524) less than the number of data used by random sampling and the second-best method, respectively.
This illustrates that DOKT can improve the quality of large-scale datasets, benefitting from the traceback to select more diverse samples.

\subsubsection{Performance on iNaturalist}
iNaturalist is a large-scale dataset with an imbalanced label distribution.
The experimental results are shown in Fig.~\ref{f2} (d).
First, DOKT outperforms the other AL methods by margins of 4.25\% over random sampling and 1.89\% over the second-best method.
This finding verifies the scalability and robustness of DOKT for cases with different data scales and label imbalances.
Second, as shown in Table~\ref{tabnum}, DOKT achieves competitive performance with only 22.5\% (105,003) of the total samples, which are 20.0\% (26,250) and 12.8\% (15,452) less than those of random sampling and the second-best method, respectively.
This indicates that DOKT can overcome the label imbalance problem to efficiently implement AL annotation.
This advantage derives from the ability of the downstream and pretext spaces to trace data interactions in the downstream domain knowledge to balance labeled datasets.

\subsection{Experimental Results on Semantic Segmentation} 
We also evaluate the performance of AL methods on semantic segmentation, which is more challenging and practical.
\noindent \textbf{Datasets.}
For the segmentation task, we test on Cityscapes~\cite{cordts2016cityscapes} and BraTS~\cite{menze2014multimodal}.
Cityscapes consists of 5,000 images of urban street scenes, and we convert it into 19 classes.
The BraTS 2018 dataset is a medical dataset that contains preoperative 3D MRI scans of 285 patients with brain tumors.
Following~\cite{gong2021ensemble}, we use 242 subjects for training and 43 for testing.

\noindent \textbf{Compared methods.} 
We compare DOKT with several state-of-the-art AL approaches that are available for this task, including LL~\cite{yoo2019active}, TA-VAAL~\cite{kim2021task}, PT4AL~\cite{yi2022pt4al}, and ALFA~\cite{parvaneh2022active}.
We also introduce the random sampling method as a baseline.

\noindent \textbf{Performance measurement.}
Following the settings of previous AL methods, we select 10\% of the available random data as the initial labeled pool.
We rigorously report the mean performance achieved across 5 trials with different random initializations.
In each iteration, 5\% of the samples are selected to be labeled, and the label budget is 40\% of the dataset.
The target model used for the performance evaluation is PSPNet~\cite{zhao2017pyramid}.

Figs.~\ref{f3} (a) and (b) show the results obtained on Cityscapes and BraTS, which involve traffic and medical data from real-world applications, respectively.
First, DOKT outperforms the other state-of-the-art methods in all iterations on both datasets.
Specifically, ALFA~\cite{parvaneh2022active} and PT4AL~\cite{yi2022pt4al} perform worse because their pixelwise outputs have difficulty estimating uncertainty via random mixing and the pretext loss.
In contrast, DOKT utilizes perceptual perturbation to enable domain mixing for complex data formats and traces downstream domain knowledge to embed high-level features for selection purposes.
With their help, DOKT can efficiently extract more informative samples than other AL methods in semantic segmentation.
Second, the gaps between DOKT and the other AL methods on segmentation are larger than those observed on classification datasets, benefitting from the ability of traceback diversity and domain uncertainty to optimize the distributions of both traffic and medical data.
This demonstrates that DOKT performs better than other methods in complex real-world applications.

\begin{table*}
	\caption{Experiment result for ablation study on SVHN dataset.}
	\label{tab2}
	\begin{center}
		\begin{tabular}{ p{2.9cm}| p{0.73cm}<{\centering} | p{0.73cm}<{\centering} | p{0.73cm}<{\centering} | p{0.73cm}<{\centering} | p{0.73cm}<{\centering} | p{0.73cm}<{\centering} | p{0.73cm}<{\centering} | p{0.73cm}<{\centering} | p{0.73cm}<{\centering} | p{0.73cm}<{\centering}}
			
			\hline
			\multirow{2}{*}{Method} & \multicolumn{10}{c}{Sampling ratio(\%)} \\
			\cline{2-11} &2&4&6&8&10&12&14&16&18&20 \\			
			\hline
			Coreset & 5.95  & \cellcolor[rgb]{ .776,  .878,  .706}20.14 & \cellcolor[rgb]{ .886,  .937,  .855}38.99 & \cellcolor[rgb]{ .886,  .937,  .855}54.62 & \cellcolor[rgb]{ .953,  .976,  .929}59.40 & \cellcolor[rgb]{ .953,  .976,  .929}64.19 & \cellcolor[rgb]{ .953,  .976,  .929}73.62 & \cellcolor[rgb]{ .886,  .937,  .855}79.63 & \cellcolor[rgb]{ .886,  .937,  .855}82.47 & \cellcolor[rgb]{ .886,  .937,  .855}83.05 \\
			MAE-Coreset & 5.95  & \cellcolor[rgb]{ .886,  .937,  .855}19.84  & \cellcolor[rgb]{ .776,  .878,  .706}39.41 & \cellcolor[rgb]{ .776,  .878,  .706}56.62 & \cellcolor[rgb]{ .776,  .878,  .706}63.00 & \cellcolor[rgb]{ .776,  .878,  .706}69.19 & \cellcolor[rgb]{ .886,  .937,  .855}76.02 & \cellcolor[rgb]{ .776,  .878,  .706}80.13 & \cellcolor[rgb]{ .776,  .878,  .706}83.07 & \cellcolor[rgb]{ .886,  .937,  .855}83.95 \\
			DOKT-T\&M & 5.95  & \cellcolor[rgb]{ .953,  .976,  .929}18.13 & \cellcolor[rgb]{ .953,  .976,  .929}35.04 & \cellcolor[rgb]{ .953,  .976,  .929}52.63 & \cellcolor[rgb]{ .953,  .976,  .929}59.08 & \cellcolor[rgb]{ .953,  .976,  .929}65.21 & \cellcolor[rgb]{ .953,  .976,  .929}72.64 & \cellcolor[rgb]{ .953,  .976,  .929}77.43 & \cellcolor[rgb]{ .953,  .976,  .929}80.59 & \cellcolor[rgb]{ .953,  .976,  .929}81.31 \\
			DOKT-M & 5.95  & \cellcolor[rgb]{ .953,  .976,  .929}18.33 & \cellcolor[rgb]{ .886,  .937,  .855}38.64 & \cellcolor[rgb]{ .886,  .937,  .855}55.25 & \cellcolor[rgb]{ .776,  .878,  .706}63.41 & \cellcolor[rgb]{ .886,  .937,  .855}68.91 & \cellcolor[rgb]{ .776,  .878,  .706}77.13 & \cellcolor[rgb]{ .776,  .878,  .706}80.47 & \cellcolor[rgb]{ .886,  .937,  .855}82.47 & \cellcolor[rgb]{ .776,  .878,  .706}84.65 \\
			DOKT-T & 5.95  & \cellcolor[rgb]{ .886,  .937,  .855}18.47 & \cellcolor[rgb]{ .953,  .976,  .929}36.01 & \cellcolor[rgb]{ .953,  .976,  .929}53.25 & \cellcolor[rgb]{ .886,  .937,  .855}61.41 & \cellcolor[rgb]{ .886,  .937,  .855}67.15 & \cellcolor[rgb]{ .953,  .976,  .929}74.01 & \cellcolor[rgb]{ .886,  .937,  .855}78.86 & \cellcolor[rgb]{ .953,  .976,  .929}81.23 & \cellcolor[rgb]{ .953,  .976,  .929}82.00 \\
			DOKT-LL  & 5.95  & \cellcolor[rgb]{ .953,  .976,  .929}17.95 & \cellcolor[rgb]{ .886,  .937,  .855}38.27 & \cellcolor[rgb]{ .886,  .937,  .855}55.03 & \cellcolor[rgb]{ .776,  .878,  .706}63.14 & \cellcolor[rgb]{ .886,  .937,  .855}68.54 & \cellcolor[rgb]{ .776,  .878,  .706}76.87 & \cellcolor[rgb]{ .776,  .878,  .706}80.22 & \cellcolor[rgb]{ .886,  .937,  .855}82.11 & \cellcolor[rgb]{ .776,  .878,  .706}84.23 \\
               DOKT-MSE  & 5.95  & \cellcolor[rgb]{ .953,  .976,  .929}17.52 & \cellcolor[rgb]{ .953,  .976,  .929}37.91 & \cellcolor[rgb]{ .953,  .976,  .929}54.18 & \cellcolor[rgb]{ .886,  .937,  .855}62.51 & \cellcolor[rgb]{ .886,  .937,  .855}68.34 & \cellcolor[rgb]{ .776,  .878,  .706}76.31 & \cellcolor[rgb]{ .886,  .937,  .855}79.80 & \cellcolor[rgb]{ .953,  .976,  .929}81.49 & \cellcolor[rgb]{ .886,  .937,  .855}83.54 \\
               DOKT  & 5.95  & \cellcolor[rgb]{ .663,  .816,  .557}21.57  & \cellcolor[rgb]{ .663,  .816,  .557}40.53  & \cellcolor[rgb]{ .663,  .816,  .557}57.69  & \cellcolor[rgb]{ .663,  .816,  .557}66.26  & \cellcolor[rgb]{ .663,  .816,  .557}71.38  & \cellcolor[rgb]{ .663,  .816,  .557}78.85  & \cellcolor[rgb]{ .663,  .816,  .557}83.34  & \cellcolor[rgb]{ .663,  .816,  .557}86.09  & \cellcolor[rgb]{ .663,  .816,  .557}86.99  \\
			\hline
		\end{tabular}
	\end{center}
\end{table*}

\subsection{Experimental Results on Image Captioning} 
We evaluate DOKT on image captioning datasets to verify its effectiveness in multimodal tasks.

\noindent \textbf{Dataset.} 
MS-COCO~\cite{LinMBHPRDZ14} is the most popular image captioning dataset.
As annotated captions of the official testing images are not provided, we utilize the Karpathy split~\cite{KarpathyL15} for this experiment. The Karpathy split has 113287/5000/5000 training/validation/testing images, respectively.

\noindent \textbf{Compared methods.} 
In this experiment, we compare our method with Multi-label~\cite{gal2016dropout}, SSAAL~\cite{zhang2020structural}, TA-VAAL~\cite{kim2021task}, PT4AL~\cite{yi2022pt4al}, and random sampling.

\noindent \textbf{Performance measurement.}
Following the settings of the previously developed AL methods, we initialize the labeled pool with 2\% of the samples and label 1\% of the samples in each iteration until the labeled ratio reaches 9\%.
To objectively evaluate the performance of the tested methods, we use four standard automatic evaluation metrics: BLEU4~\cite{papineni2002bleu}, METEOR~\cite{denkowski2014meteor}, ROUGE-L~\cite{Lin2004ROUGE}, and CIDEr~\cite{vedantam2015cider}. The obtained METEOR and ROUGE-L results are shown in the supplementary material.

Figs.~\ref{f3} (c) and (d) show the results using the BLEU4 and CIDEr metrics, respectively.
First, DOKT evidently achieves the highest metric scores in each iteration.
This indicates that DOKT can select more informative cross-modal samples.
Furthermore, DOKT outperforms the second-best method (SSAAL)~\cite{zhang2020structural}, with average margins of 1.31 and 2.71 in terms of BLEU4 and CIDEr, respectively, verifying the superiority of DOKT.
Second, with 9\% of the labeled data, DOKT achieves scores of 32.68 (BLEU4) and 99.18 (CIDEr), which are close to those of the image captioning model with 100\% of the labeled data.
This competitive performance benefits from the use of a traceback module that integrates low- and high-level features to improve the sampling quality of the proposed method.
Third, PT4AL~\cite{yi2022pt4al} and DOKT both utilize pretext guidance, while DOKT integrates downstream annotations to trace the data relationships between the low- and high-level spaces.
As a result, DOKT can select the most instructive samples with diverse descriptions from the unlabeled pool.

\subsection{Ablation Study}
To deeply analyze the effectiveness of key modules in DOKT, we perform the ablation study.
First, we conduct the ablation to directly use the pre-trained model as follows:
(1) Coreset: Coreset method. 
(2) MAE-Coreset: Coreset~\cite{sener2017active} method with data representations of the pre-trained model(MAE~\cite{he2022masked}). 
Further, we conduct the ablation on the traceback diversity, domain uncertainty and the ranking loss as follows:
(3) DOKT-T\&M: DOKT without both of the above. 
(4) DOKT-M: DOKT without domain mixing. 
(5) DOKT-T: DOKT without traceback module. 
(6) DOKT-LL: DOKT  using  LL loss. 
(7) DOKT-MSE: DOKT using  MSE loss. 
(8) DOKT: the proposed method. 

\begin{figure*}
	\centering
	\subfloat[{\footnotesize  LL}]{\includegraphics[width=0.23\linewidth]{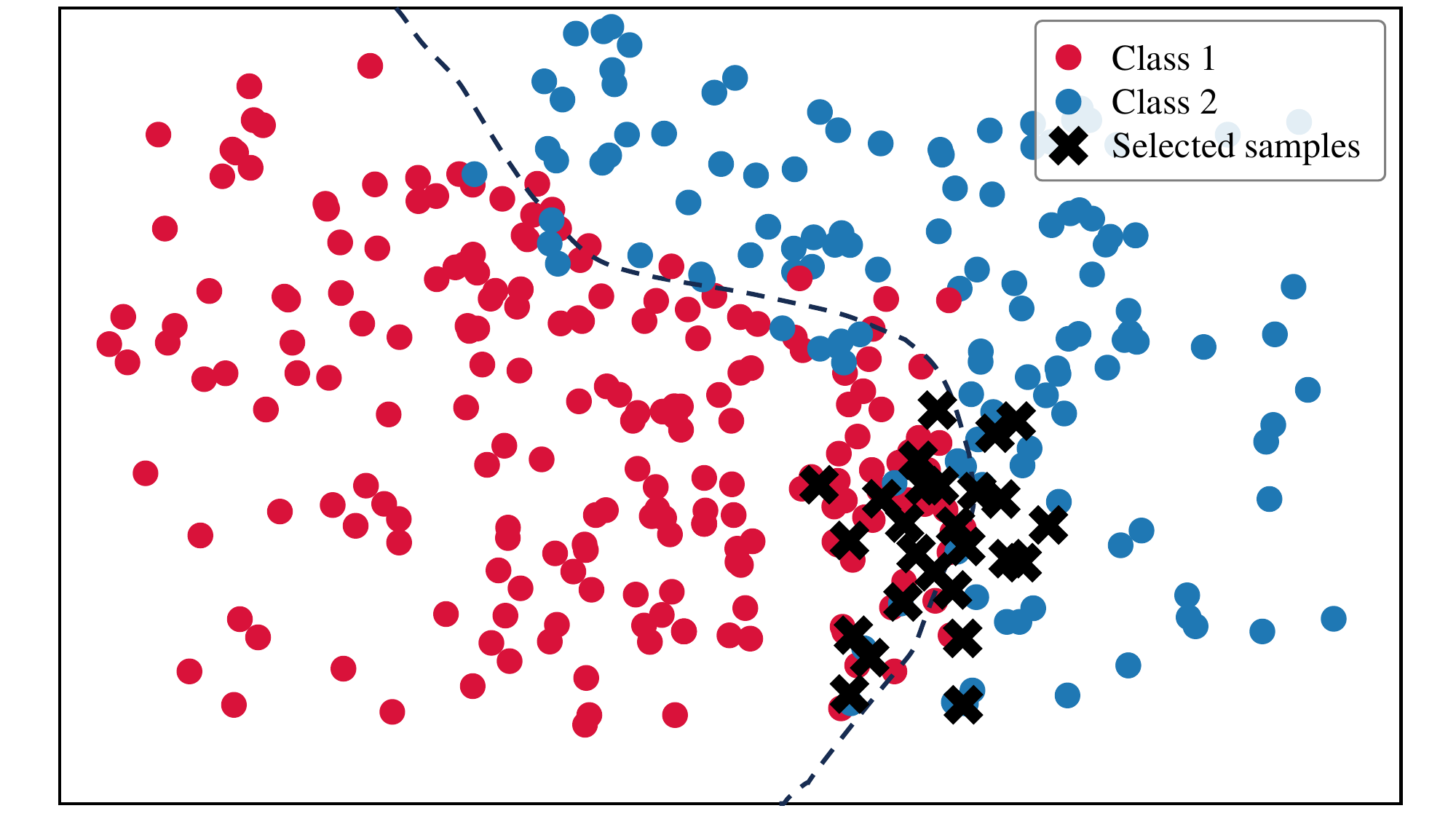}} 
	\subfloat[{\footnotesize  ALFA}]{\includegraphics[width=0.23\linewidth]{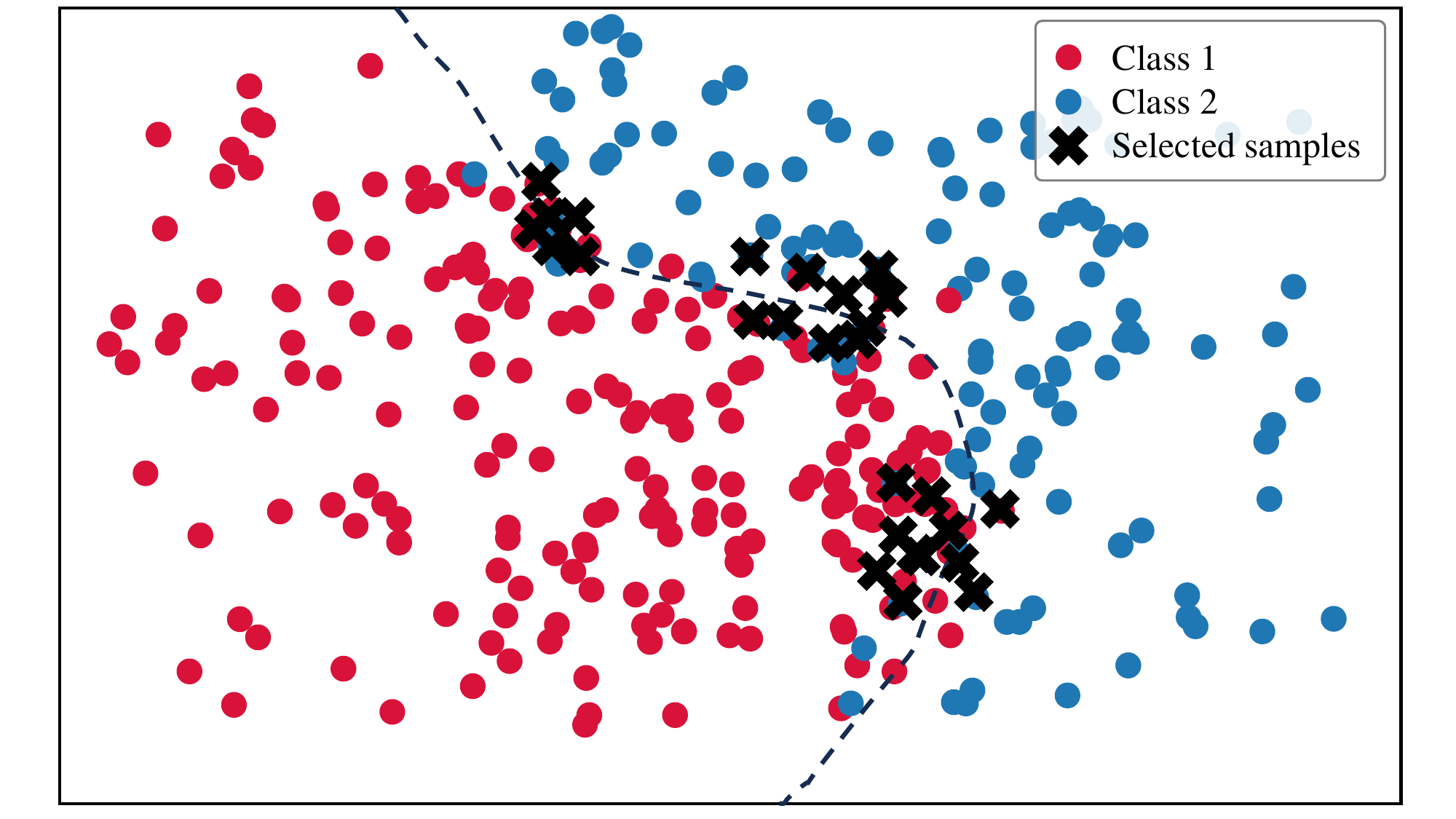}} 
	\subfloat[{\footnotesize  PT4AL}]{\includegraphics[width=0.23\linewidth]{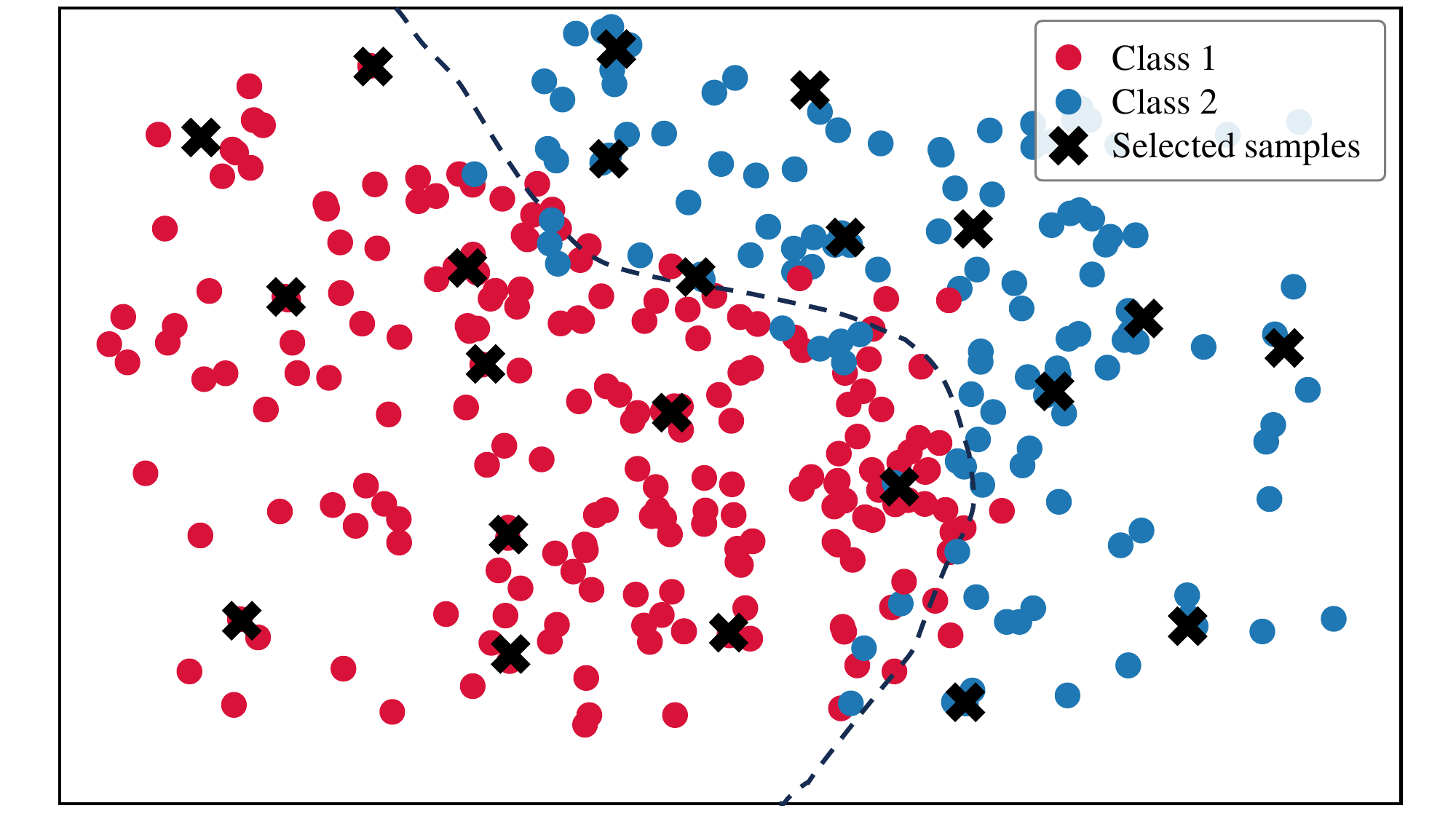}}
	\subfloat[{\footnotesize BAL}]{\includegraphics[width=0.23\linewidth]{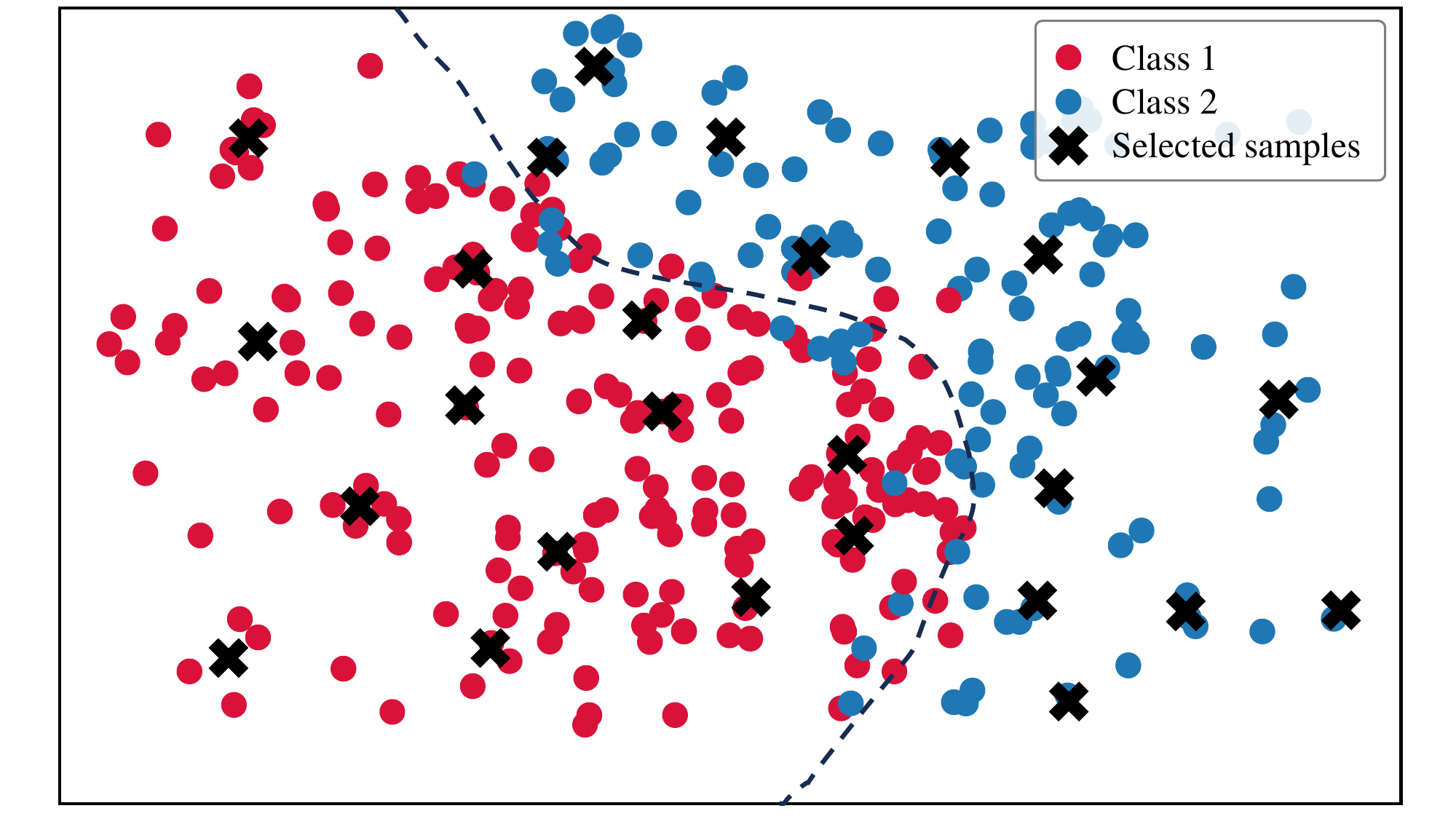}} 
 
	\subfloat[{\footnotesize  DOKT}]{\includegraphics[width=0.23\linewidth]{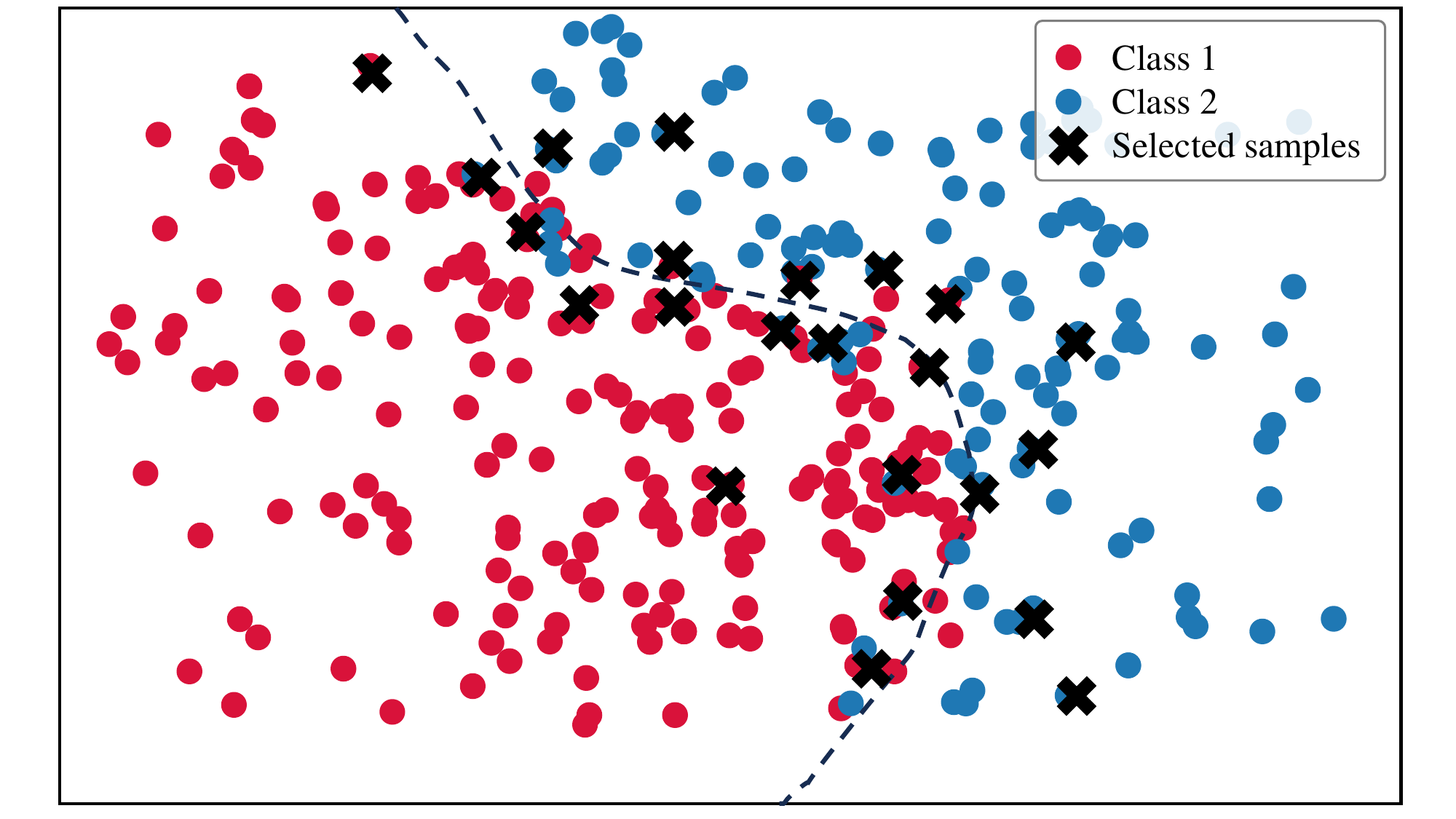}} 
 \subfloat[{\footnotesize  DOKT ($\iota=L / 2$ )}]{\includegraphics[width=0.23\linewidth]{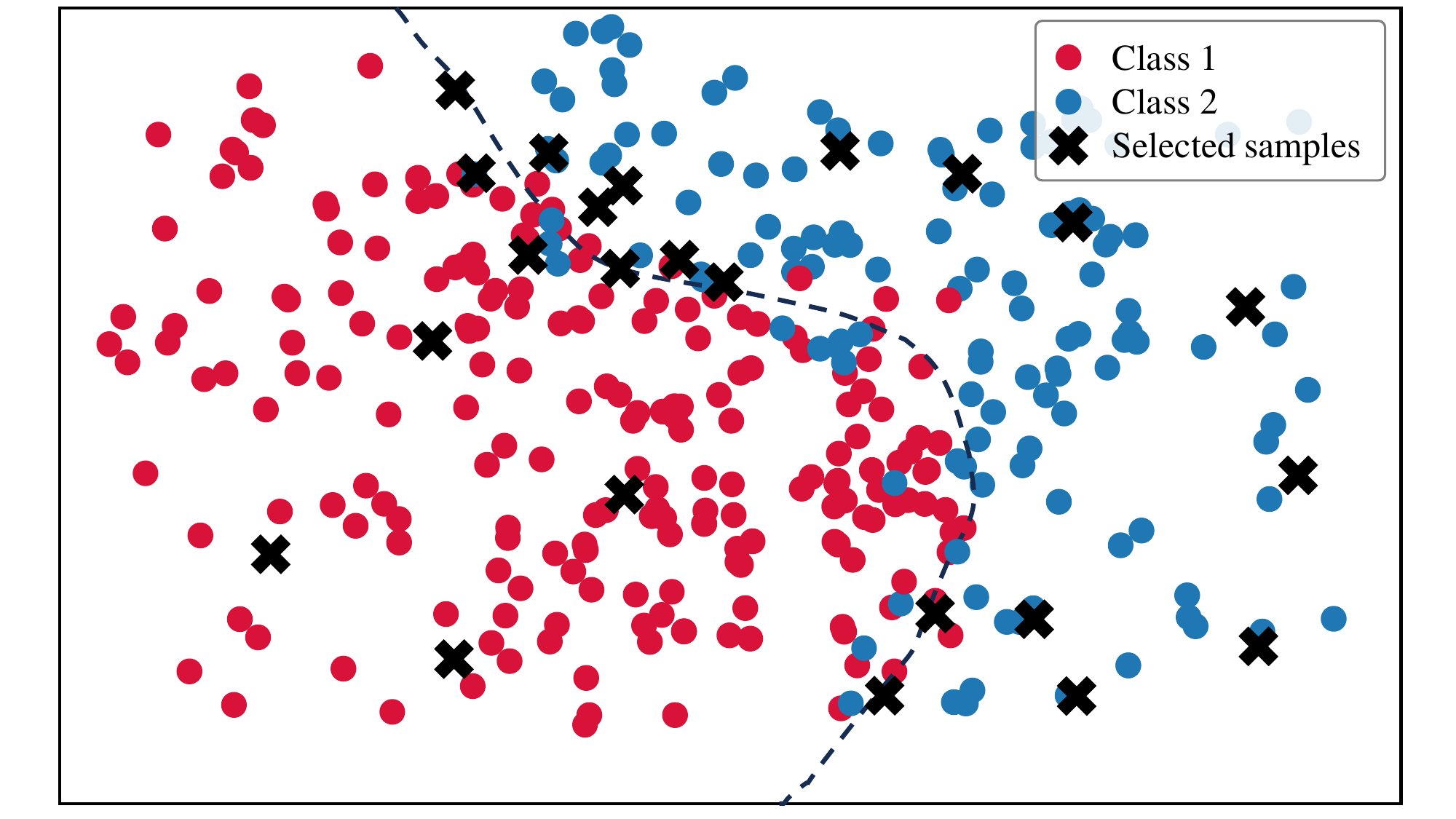}}
	\subfloat[{\footnotesize DOKT (LL loss)}]{\includegraphics[width=0.23\linewidth]{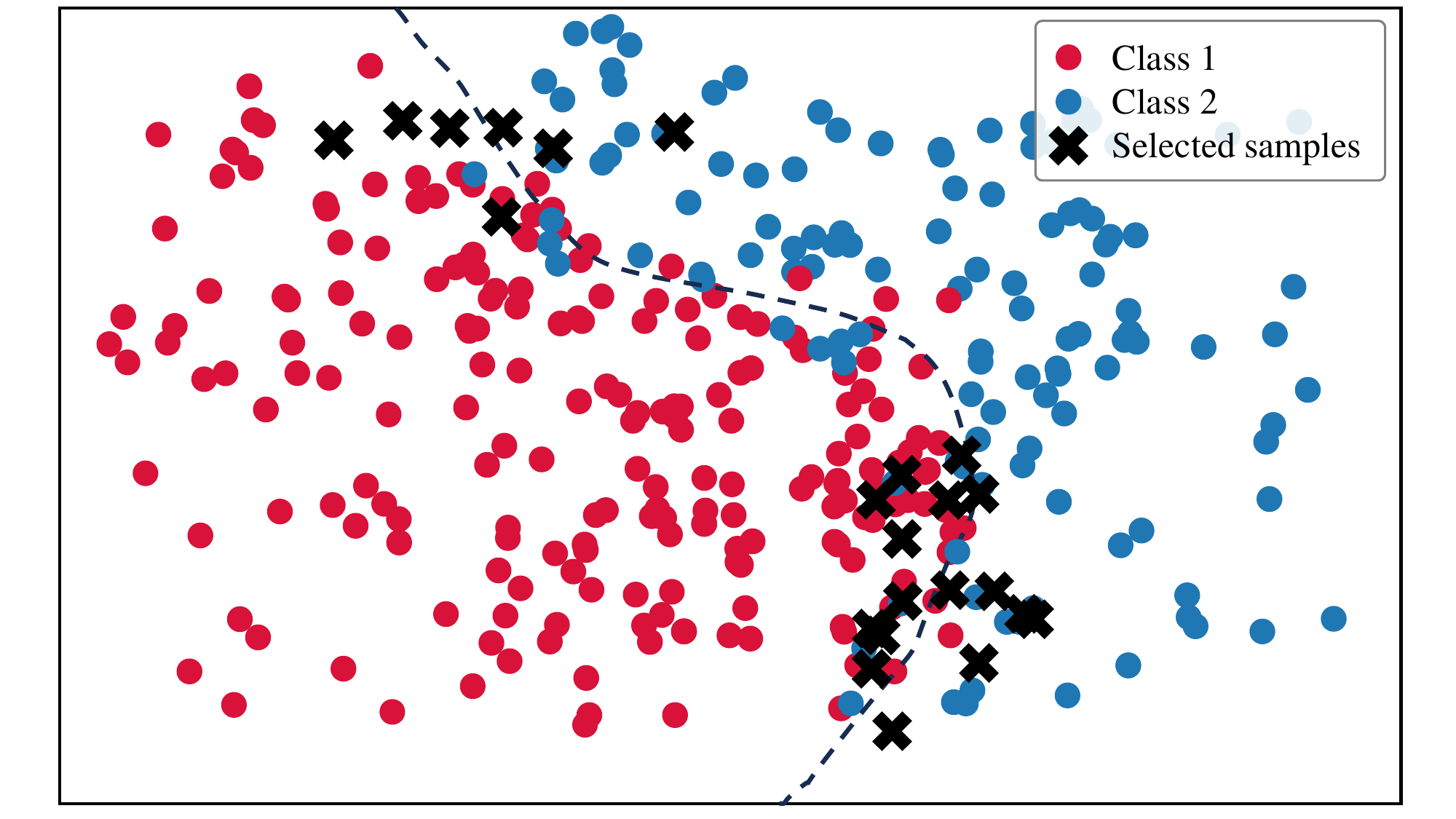}} 
	\subfloat[{\footnotesize  DOKT (MSE loss)}]{\includegraphics[width=0.23\linewidth]{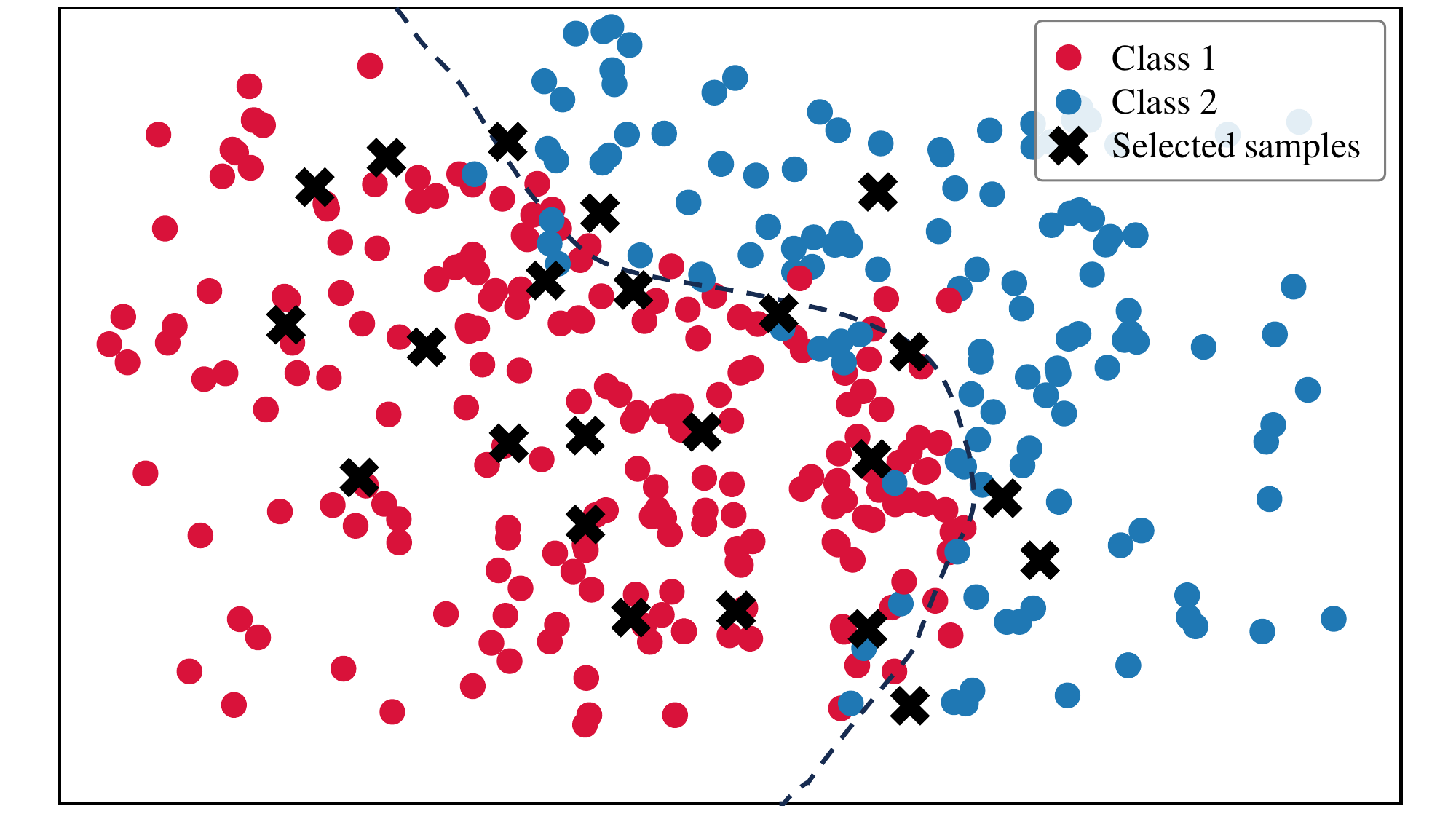}} 
 
	\caption{Visualization of AL sampling results on CIFAR-10. Red and blue dots indicate samples assigned to classes 0 and 1, respectively. The decision boundary of the oracle annotation is shown as a black dashed line. (a)-(e) visualize the sampling results of different AL methods (f) visualizes the result of DOKT with fixed $\iota=L / 2$. (g)-(h) visualize the results of DOKT with different loss functions. }
	\label{img3}
\end{figure*}

The ablation is shown in Table~\ref{tab2}. 
First, we can observe that DOKT consistently outperforms DOKT-D and DOKT-U by a significant improvement, which verifies the superiority of DOKT.
Second, DOKT-D and DOKT-U both perform better than DOKT-T\&M.
The above phenomenon illustrates that both the traceback diversity and the domain mixing augmentation help to improve the AL performance, where the traceback module covers the whole data distribution and the domain mixing alleviates the annotation insufficiency for uncertainty estimation.
Third, MAE-Coreset performs better than Coreset, and this verifies that the high-quality representation of pre-training can boost AL selection.
However, MAE-Coreset still obviously underperforms DOKT.
This indicates that the rudimentary manner of using pre-training in AL methods only brings limited improvement due to the lack of domain knowledge.
The comparison of DOKT and MAE-Coreset demonstrates that DOKT effectively leverages the pre-trained guidance by bridging it with domain knowledge. 
Fourth, both DOKT-LL and DOKT-MSE perform not well, and they even perform worse than DOKT-M.
Since the MSE prediction is unstable and the LL loss cannot constrain the value range, it not suitable to combine them with prediction divergence. 
The large margins between DOKT and the two ablation groups demonstrate that the ranking loss can provide a robust uncertainty estimation to improve the quality of labeled datasets.

\subsection{Visualization of AL Sampling}
To analyze the data distribution of AL sampling results, we show the t-SNE~\cite{vandermaaten08a} visualization in  Fig.~\ref{img3}.
In the plots, we visualize samples from two categories in CIFAR-10 and display the decision boundary for better analysis.
(a)-(e) show the results of different AL methods; (f) shows the result of DOKT with fixed $\iota$. (g)-(h) show the results of DOKT with different loss functions.
For LL~\cite{yoo2019active} in Fig.~\ref{img3}(a), the learning loss method tends to select difficult samples but lacks information about global distribution.
Thus, we can find that its selected samples are located close to the decision boundary but are clustered in a narrow region, which leads to sampling overlapping and inefficiency.
Besides, ALFA~\cite{parvaneh2022active} in Fig.~\ref{img3}(b) estimates the data uncertainty and inconsistency.
Although random mixing in ALFA helps to find samples that spread spatially, the  random process  is uncontrollable and  results in the local cluster.
For PT4AL~\cite{yi2022pt4al} in Fig.~\ref{img3}(c), since it estimates sampling value with pretext loss that only involves low-level perception, the selection is diverse but ignores the uncertain samples near the boundary.
The same problem occurs for BAL~\cite{zhang2023scalable} in Fig.~\ref{img3}(d).
BAL only focuses on balancing label distribution, so that it allocates the annotation equally to classes 1 and 2 while the selection is far from the boundary and not instructive.
For DOKT in Fig.~\ref{img3}(e), we can observe that the selected samples are all near the decision boundary and there is no sampling cluster, which is much more efficient.
This demonstrates that DOKT can select instructive and diverse samples near the decision boundary, where the traceback module traces data interaction of pre-training guidance and domain knowledge to improve the quality of labeled datasets.
For fixed $\iota$ in Fig.~\ref{img3}(f), we can observe that the fixed $\iota$ cannot select suitable neighbors to explore the data relationships, which verifies that the adaptive $\iota$ helps to estimate the data diversity in two spaces for better AL sampling.
For Fig.~\ref{img3}(g), since LL loss cannot constrain the value range, it is impossible to balance the diversity and uncertainty score so that there is obvious sampling overlapping.
For Fig.~\ref{img3}(h), selected samples are not near the boundary due to the difficult convergence of MSE. The results demonstrate that the ranking loss can help DOKT to select instructive samples.

\section{Conclusion and Future work}
\label{sec:5}
In this paper, we analyze the active learning (AL) mechanism and propose a downstream-pretext domain knowledge traceback model (DOKT)  that traces two spaces of pre-training guidance and domain knowledge to explore data interaction for constructing high-quality labeled datasets. The model consists of a traceback diversity indicator that explores data diversity in low-level and high-level spaces, and a domain-based uncertainty estimator that learns to predict uncertainty scores by perceptual perturbing for domain mixing samples. The experiments demonstrate that DOKT boosts model training with significant label reduction and generalizes for various application scenarios. We hope DOKT will spawn ideas for AL by using pretext training and downstream knowledge.
			
Although DOKT can utilize a pre-trained model, it can only handle features and cannot analyze the large parameters in the pre-trained model for uncertainty estimation. This makes DOKT inefficient for constructing high-quality labeled data for transferring large models to specific downstream tasks. In the future, we will extend the proposed DOKT algorithm to label important samples for transferring large pre-trained models. Since large pre-trained models still require downstream labeled data for generalizing to a specific task, the use of AL in pre-trained models is desirable for related research. The main challenges are (1) the complexity of model structures and (2) the wide range of domain knowledge, which make it difficult for traditional AL methods to estimate model uncertainty and adapt to downstream tasks. In view of these limitations, we will attempt to optimize the AL process and design new strategies to improve our method for use with large pre-trained models.

\bibliographystyle{IEEEtran}
\bibliography{egbib}


\vfill

\end{document}